\pdfoutput=1

\documentclass[11pt]{article}

\usepackage[final]{acl}
\usepackage{graphicx}
\usepackage{times}
\usepackage{latexsym}
\usepackage{float}  
\usepackage{makecell} 
\usepackage{placeins}  
\usepackage{stfloats}

\usepackage[T1]{fontenc}

\usepackage[utf8]{inputenc}

\usepackage{microtype}

\usepackage{inconsolata}

\usepackage{graphicx}

\usepackage{ragged2e}
\usepackage{hyperref}
\usepackage{multirow}
\usepackage{graphicx}
\usepackage{xspace}
\usepackage{color}
\usepackage{colortbl}
\usepackage{tablefootnote}
\usepackage{pifont}
\usepackage{makecell}
\usepackage{framed}
\usepackage{mdframed}
\usepackage{subfigure}
\usepackage{caption}
\usepackage{longtable}
\usepackage{float}
\usepackage{booktabs}
\usepackage{arydshln} 

\usepackage{amsfonts}
\usepackage{pgfplots}
\usepackage{pgfplotstable}
\usepackage{booktabs}

\newcolumntype{H}{>{\setbox0=\hbox\bgroup}c<{\egroup}@{}}

\newcommand{\ie}{\emph{i.e.,}\xspace}

\newcommand{\eg}{\emph{e.g.,}\xspace}

\newcommand{\ignore}[1]{}

\usepackage{tcolorbox}
\usepackage{algpseudocode}
\usepackage{amsthm}
\usepackage{amsmath}
\usepackage{tikz}

\usepackage{amsmath}

\definecolor{TableRowColor}{HTML}{F3E6FA}
\definecolor{SoftBlue}{RGB}{135, 206, 250}
\definecolor{SoftBlue}{RGB}{135, 206, 250}
\definecolor{SoftOrange}{RGB}{255, 224, 178}
\definecolor{SoftGreen}{RGB}{144, 238, 144}
\definecolor{CorrectGreen}{RGB}{76, 175, 80}
\definecolor{ErrorRed}{RGB}{211, 47, 47}
\definecolor{TableRowColor}{HTML}{F3E6FA}
\definecolor{TextBackColor}{HTML}{D0DFE5}

\title{LiveLongBench: Tackling Long-Context Understanding \\ for Spoken Texts from Live Streams}

\author{
	Yongxuan Wu$^{1}$\thanks{$\ $ Authors contributed equally.}, \  
	Runyu Chen$^{1*}$, \
	Peiyu Liu$^{1}$\thanks{$\ $ Corresponding author.}, \  
	Hongjin Qian$^{2}$\ 
	\\
	$^1$University of International Business and Economics\\
	$^2$Beijing Academy of Artificial Intelligence\\
	\tt{wuyongxuanyy@gmail.com, ry.chen@uibe.edu.cn} \\ \tt{liupeiyustu@163.com, chienqhj@gmail.com}\\ 
}

\begin{document}
\maketitle
\begin{abstract}
Long-context understanding poses significant challenges in natural language processing, particularly for real-world dialogues characterized by speech-based elements, high redundancy, and uneven information density. Although large language models (LLMs) achieve impressive results on existing benchmarks, these datasets fail to reflect the complexities of such texts, limiting their applicability to practical scenarios. 
To bridge this gap, we construct the first spoken long-text dataset, derived from live streams, designed to reflect the redundancy-rich and conversational nature of real-world scenarios. 
We construct tasks in three categories: retrieval-dependent, reasoning-dependent, and hybrid. We then evaluate both popular LLMs and specialized methods to assess their ability to understand long-contexts in these tasks.
Our results show that current methods exhibit strong task-specific preferences and perform poorly on highly redundant inputs, with no single method consistently outperforming others.
We propose a new baseline that better handles redundancy in spoken text and achieves strong performance across tasks. Our findings highlight key limitations of current methods and suggest future directions for improving long-context understanding.
Finally, our benchmark fills a gap in evaluating long-context spoken language understanding and provides a practical foundation for developing real-world e-commerce systems.
The code and benchmark are available at \url{https://github.com/Yarayx/livelongbench}.

\end{abstract}

\section{Introduction}
\label{sec-intro}
Spoken texts, prevalent in scenarios such as dialogues and live streams, are becoming increasingly common as conversational AI and real-time communication continue to expand. 
Existing studies have demonstrated that spoken text exhibits unique linguistic properties~\cite{eisenstein2013what}, particularly~\emph{high redundancy} characterized by repetitive phrases and filler words.
This redundancy imposes significant computational challenges, including increased processing overhead and difficulties in semantic understanding.
While advanced LLMs support long context lengths~\cite{touvron2023llama} and current Key-Value~(KV) cache compression methods~\cite{liu2024kivi,jiang2024minference,pan2024llmlingua} have been designed for written texts, their ability to handle the unique redundancy patterns of spoken texts remains unexplored. 
This gap underscores the need for specialized approaches tailored to the characteristics of spoken language.

Generally, long contexts pose challenges for both understanding and computation. LLMs often struggle with lengthy texts, such as the~\emph{lost in the middle} phenomenon~\cite{liu2024lost}. However, existing benchmarks~\cite{longbench,inftybench} for long-context understanding predominantly focus on written texts, neglecting the informal characteristics of spoken language. 
Beyond these understanding challenges, LLMs often waste computation on filler words (\eg ``um'', ``uh'') and other redundant content. These tokens contribute little semantic value but significantly expand the KV cache, increasing memory and latency costs. This motivates the need for more aggressive and selective context compression strategies. To explore this, we raise two central questions:

~\emph{Question (1): Can base models effectively process long spoken texts with informal language characteristics?}

~\emph{Question (2): Can existing methods achieve higher compression rates, for example, through the combination of multiple techniques?}

\begin{table*}[t]
    \centering
    \small
    \begin{tabular}{l c c c c c c c c}
        \toprule 
        & \multicolumn{2}{c}{\textbf{Response Type}} & \multicolumn{2}{c}{\textbf{Attention Span}} & \multicolumn{2}{c}{\textbf{Language Style}} \\[0.3em]
        \textbf{Dataset} & Closed & Open & Global & Semantic & Spoken Texts & Languages & \textbf{Avg. Words} \\
        \midrule
        LongBench  & \checkmark & \checkmark & \checkmark &  &  & En.\&Zh. & \textasciitilde 13k \\ 
        $\infty$Bench  & \checkmark &  &  &  & \checkmark  & En. & \textasciitilde 300k \\ 
        Loong  &  & \checkmark  &  & \checkmark &   & En.\&Zh. & \textasciitilde 110k \\ 
        Marathon  & \checkmark &  &  &  &   & En. & \textasciitilde 163k \\ 
        L-Eval  & \checkmark & \checkmark & \checkmark & \checkmark & \checkmark  & En.\&Zh. & 3k - 62k \\
        M4LE  & \checkmark & \checkmark & \checkmark & \checkmark &   & En.\&Zh. & \textasciitilde 4k \\
        TCELongBench  & \checkmark & \checkmark  &  & \checkmark  & \checkmark & En. & \textasciitilde 18k \\ 
        FinTextQA  &  & \checkmark  & \checkmark & \checkmark &   & En. & \textasciitilde 19k \\ \midrule
        \textbf{LiveLongBench}  & \checkmark & \checkmark & \checkmark & \checkmark & \checkmark  & En.\&Zh. & \textasciitilde 97k \\ 
        \bottomrule    
    \end{tabular}
    \caption{Comparison of Different Long-context Benchmark Datasets.}
    \label{tab:long_benchmarks}
\end{table*}

To assess how well existing LLMs and context compression methods handle long-form spoken texts, we construct a new benchmark, \textbf{LiveLongBench}, tailored to the challenges of this setting.
As a core component, we construct a novel dataset recorded from live streams, featuring both Chinese and English, with sequences averaging approximately 97K tokens. To tackle the first question, we follow the study~\cite{wang2024leave} and design 9 distinct tasks that fall into three categories: \emph{retrieval}, \emph{reasoning}, and \emph{hybrid tasks}.
For each category, we synthesize multiple questions to evaluate various model capabilities, covering both open-domain and closed-domain settings to assess knowledge recall and generalization. 
Furthermore, according to the findings in~\cite{m4le}, the critical information required for task completion in long sequences can be categorized by relevance into single-span, multiple-span, or global. 
Specifically, global tasks involve reasoning over the entire context and can be viewed as generalizing over other span types.
As an important extension, we introduce {semantic multi-span}, which focuses on semantically distributed spans rather than merely structurally separated paragraphs. This task type can be seen as an advanced form of multi-span reasoning, requiring models to integrate and infer over multiple conceptually related but dispersed segments.
Built upon these designs, LiveLongBench offers a thorough evaluation framework for long-context understanding in spoken language, which remains underexplored in existing benchmarks.



To address the second question, we first evaluate individual KV cache compression methods, including KIVI~\cite{liu2024kivi}, MInference~\cite{jiang2024minference}, and Lingua~\cite{pan2024llmlingua}. 
Interestingly, we discover that \textbf{certain method combinations can lead to improved performance while reducing memory consumption, outperforming individual approaches}. For example, using \emph{Minference+Lingua-4x} outperforms each single method, while using \emph{KIVI-4bit+Minference+Lingua-2x} achieves the lowest memory usage and still surpasses individual approaches such as KIVI or M-Inference. 
To further balance memory efficiency and performance, we apply a Data Envelopment Analysis (DEA) framework to evaluate the cost-effectiveness of all method combinations.  
The resulting ranking offers a practical reference for selecting the optimal combination under different deployment constraints.

Our contributions are summarized as follows:

$\bullet$~We construct and release LiveLongBench, the first bilingual benchmark derived from live-streaming spoken texts, designed to evaluate long-context understanding and reasoning, with sequences averaging approximately 97K tokens.

$\bullet$~We systematically evaluate current LLMs, uncovering significant performance degradation when processing lengthy spoken contexts and highlighting the unique challenges posed by informal language patterns.

$\bullet$~We propose a hybrid KV cache compression strategy, which combines multiple compression methods and achieves superior performance-memory trade-offs, as identified through a comprehensive DEA-based efficiency analysis.

\section{Related Work}

\paragraph{Long-context Understanding Benchmarks.}
Numerous benchmarks have been developed to evaluate long-text understanding, predominantly focusing on formal, written texts. Datasets such as~\cite{TCELongBench,loong} emphasize structured, coherent, and information-dense content, while tasks like document summarization, information retrieval, and long-form question answering have been extensively studied using datasets such as NarrativeQA~\cite{kovcisky2018narrativeqa}, MultiNews~\cite{fabbri2019multi}, and SQuAD 2.0~\cite{sulem2021we}. Although these benchmarks have driven progress in long-text understanding, their reliance on formal language overlooks the challenges posed by spoken language—characterized by disfluencies, redundancy, and variability—which leads to models that often struggle with real-world applications such as live-stream transcripts and conversational logs.

\paragraph{Conversational and Spoken Text Processing.}
Research in conversational text processing, especially within dialogue systems and ASR, has produced benchmarks such as DailyDialog~\cite{li2017dailydialog}, PersonaChat~\cite{zhang2018personalizing}, and DSTC~\cite{DSTC}.
These datasets typically consist of short, goal-oriented dialogues with minimal noise and redundancy, but they fail to capture the complexity of real-world speech.
{Traditional Spoken Language Understanding~(SLU) datasets, such as ATIS~\cite{atis} and SNIPS~\cite{snips}, while widely used, typically contain highly structured, domain-specific content (e.g., travel or SMS) with limited contextual scope and minimal redundancy.} 
In contrast, corpora like Switchboard~\cite{godfrey1992switchboard} and CallHome reflect the irregular, fragmented nature of natural spoken language, albeit in limited domains like telephony. 
Live commerce and streaming platforms are emerging as rich sources of diverse spoken data. 
However, systematic collection and analysis efforts remain limited. Recent work in video summarization (\eg TVSum~\citealp{tvsum}) and e-commerce dialogue datasets underscores the need for specialized methods, as comprehensive solutions for long-form spoken language understanding are still underdeveloped.

\paragraph{Spoken Long-Text Benchmarks: Gaps and Advances}
Existing long-text benchmarks primarily target formal written language, overlooking the redundancy, informality, and variability of spoken texts and rarely evaluating methods for redundancy reduction or long-context processing on authentic spoken data. 
As shown in Table~\ref{tab:long_benchmarks}, while LongBench~\cite{longbench} offers rich content with key evidence often confined to specific paragraphs, and benchmarks such as $\infty$Bench~\cite{inftybench}, Marathon, and Loong~\cite{loong} provide ultra-long contexts with limited question diversity, and L-Eval~\cite{Leval} and M4LE~\cite{m4le} feature varied question types over shorter contexts, and domain-specific benchmarks like TCELongBench~\cite{TCELongBench} and FinTextQA~\cite{fintextqa} target the news and finance domains, LiveLongBench preserves extensive context, offers a broader range of question types, and incorporates spoken linguistic characteristics, making it more representative of real-world spoken language.


\section{LiveLongBench}
\subsection{Basic Challenges}
\label{sec-basic}
To study long-context understanding in realistic spoken scenarios, we construct a dataset that captures the core challenges encountered in practice, particularly \emph{informal language} and \emph{high redundancy}.
Next, we will describe the construction process in detail.

\paragraph{Informal Language.}
Live-streaming e-commerce data often involves conversational speech, contributing to the informality of the language. Unlike formal text, live-stream content typically consists of short, fragmented utterances, leading to a high occurrence of syntactic reduction.
Additionally, interactive conversations with viewers frequently introduce topic drift,
where discussions shift abruptly, making it difficult for models to maintain contextual coherence. These characteristics significantly increase the complexity of document understanding compared to well-structured formal text.
\begin{figure}[!h]
\begin{tcolorbox}[colback=yellow!10,colframe=gray!85,title=Examples of the informal languages]
\small
$\vartriangleright$ \textbf{Syntactic Reduction:} \\
\emph{``Big scarf, the discount area.''}\\~\textcolor{orange}{Verbless}\\
\emph{``This place, the focus of our vision.''}\\~\textcolor{orange}{Right-Dislocation}\\
\\
$\vartriangleright$ \textbf{Topic Drift:} \\
\emph{``This handbag is made of genuine leather and comes in three colors. I bought one for my sister last week... Oh, by the way, did you see the movie I talked about yesterday?''}\\
~\textcolor{orange}{From product details to unrelated personal topics}
\end{tcolorbox}
\vspace{-1em}
\end{figure}

\paragraph{High Redundancy.}
Live-stream transcripts contain a substantial amount of filler words.
To emphasize key product features, presenters often include repetitive content 
,reiterating the same information multiple times. Furthermore, interactive dialogues introduce additional non-informative tokens, 
which inflate the overall length while lowering the density of useful information. This high redundancy poses challenges for long-context processing, requiring models to efficiently filter out noise while retaining essential details.
\begin{figure}[!h]
\begin{tcolorbox}[colback=yellow!10, colframe=gray!85, title=Examples of the redundant content]
\small
$\vartriangleright$ \textbf{Filler Words:} \\
\emph{``Um, okay, so, yeah, you know, like, I mean, actually, basically...''}\\
\\
$\vartriangleright$ \textbf{Repetitive Content:} \\
\emph{``This bag is beautiful, really beautiful, so beautiful! I mean, it’s just beautiful!''}\\
\\
$\vartriangleright$ \textbf{Non-informative Tokens:} \\
\emph{``This is really nice, you know? It’s just so good. Like, really good, you know what I mean?''}\\
\end{tcolorbox}
\end{figure}


\subsection{Dataset Collection}
To tackle the challenges of long-context compression in spoken language, we build a dataset that not only reflects real-world characteristics at scale but also enables systematic evaluation across diverse spoken tasks. 

\begin{figure}[t]
    \centering
    \includegraphics[width=0.93\linewidth]{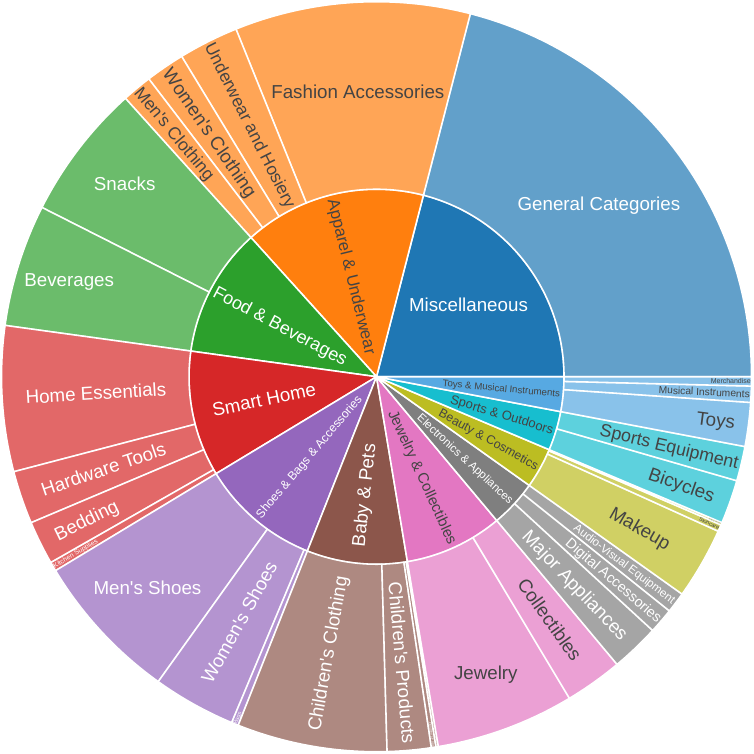}
    \caption{Distribution of Data Categories Across E-Commerce Domains}
    \label{fig:domains}
\end{figure}

\paragraph{Data Source.}
The dataset is built from {Douyin} e-commerce live streams, known for their diverse and dynamic live streaming styles.  We collected and transcribed audio from live sessions spanning 11 major product categories and 32 subcategories, including apparel, electronics, beauty, and household goods. To the end, we show the distribution of product categories in Figure~\ref{fig:domains}.

This dataset reflects how people actually talk in live-streams, making them a compelling source for building the benchmark on spoken long-context understanding.
Each document mainly contains continuous host monologues, which are characterized by informal expressions, repetitive promotional content, and frequent Q\&A exchanges as mentioned in Section~\ref{sec-basic}. This diversity ensures that this dataset accurately mirrors the linguistic challenges of real-world spoken language, providing a valuable testbed for developing compression methods tailored to the informal and redundant nature of spoken text. 
\paragraph{Processing and Structuring.}
To faithfully preserve the full characteristics of spoken language, we first transcribe the audio using the pre-trained Whisper speech-to-text model~\footnote{Whisper is pre-trained on a large-scale audio corpus.}, retaining all repetitions and filler words to maintain the authenticity of the original context.
Next, to protect user privacy and ensure the dataset is used solely for research purposes, we remove sensitive content such as personal identifiers during the processing phase.
Finally, we apply a light filtering process to eliminate extreme noise, such as repeated instances of the same sentence occurring more than ten times, while preserving the informal and redundant style of live-stream discourse. This ensures both the linguistic realism of the dataset and its usability for downstream analysis.

\subsection{Task Construction}
\begin{figure*}[t]
    \centering
    \includegraphics[width=1.0\linewidth]{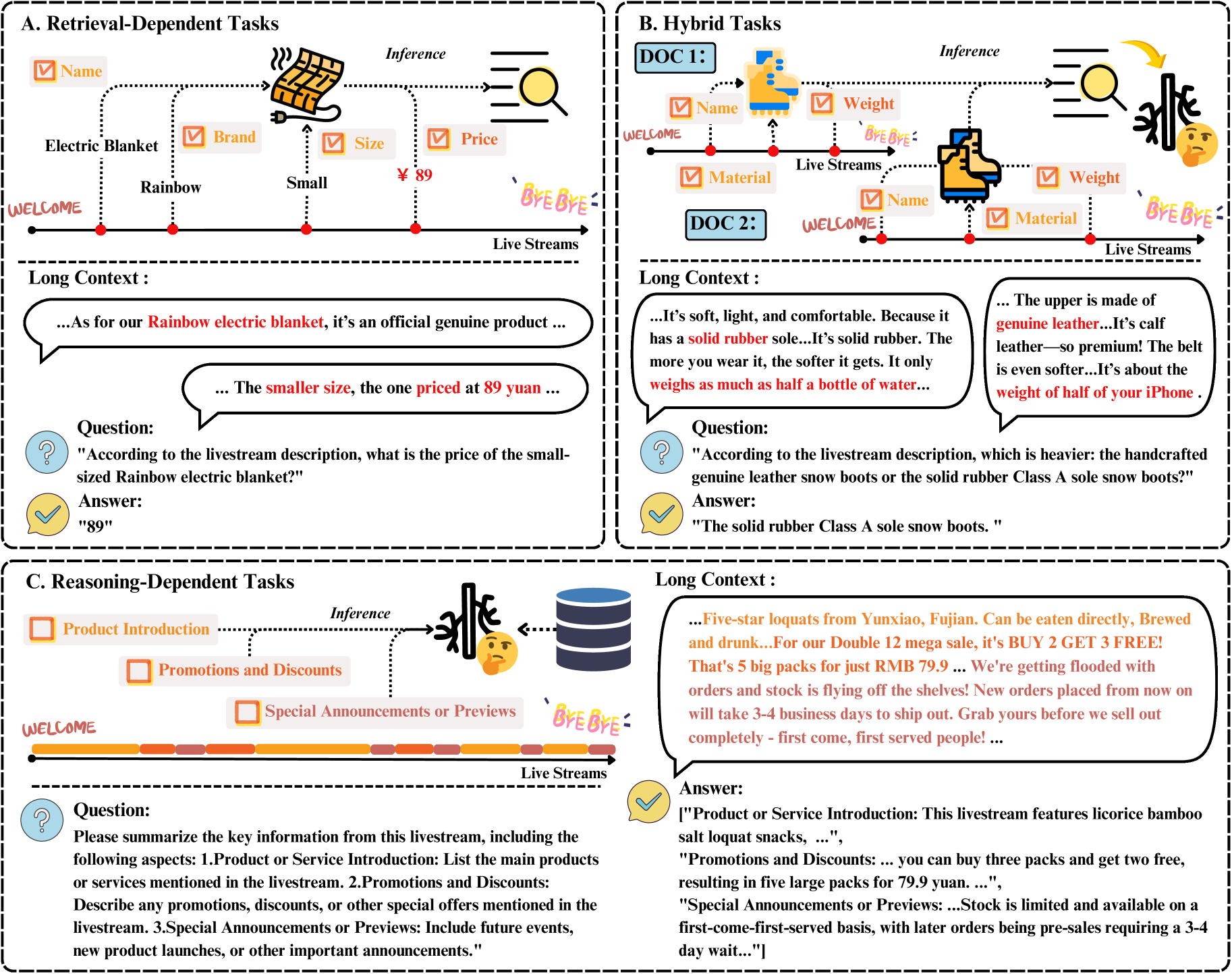}
    \caption{Showcase of Three Evaluation Tasks in LiveLongBench}
    \label{fig:tasks}
\end{figure*}

{Motivated by the study~\cite{wang2024leave}}, we define three primary task categories that align with the inherent characteristics of spoken language (see Figure~\ref{fig:tasks}): 1)~\emph{retrieval-dependent} tasks, which challenge models to extract specific information from lengthy and often redundant spoken content, 2)~\emph{reasoning-dependent} tasks, which require models to navigate informal expressions, filler words, and fragmented structures to perform complex logical inference, and 3)~\emph{hybrid tasks}, which combine both retrieval and reasoning, reflecting real-world spoken scenarios where models must identify relevant details while simultaneously reasoning over loosely structured discourse.
\paragraph{Retrieval-Dependent Tasks.}
Retrieval in this context refers to a model’s ability to locate specific information from spoken content, such as identifying product policy (\ie task ``\texttt{Policy}''), or extracting product specifications from a single document (\ie task ``\texttt{Single}''). 
These tasks may require the model to find the listed price of a product mentioned during a live-stream or to verify its attributes based on the host’s verbal descriptions.

\paragraph{Reasoning-Dependent Tasks.}
Reasoning refers to a model’s ability to infer information not explicitly mentioned in the spoken content by leveraging internal knowledge. This includes classifying a product into the correct category (\ie task ``\texttt{Class}''), which often requires external knowledge about product types and market conventions, \eg recognizing that a niche electronic device is a type of wearable.
It also includes summarizing key points from lengthy and informal conversations (\ie task ``\texttt{Summary}''), where the model must identify and synthesize essential information despite redundancy and noise,
such as distilling a coherent summary from a promotional session with repeated slogans and off-topic remarks.

\paragraph{Hybrid Tasks.}
Hybrid tasks combine both retrieval and reasoning, requiring models to first extract multiple relevant pieces of information from spoken content and then synthesize them through reasoning to form a coherent response. This includes answering questions that span multiple segments of a live-stream transcript (\ie task ``\texttt{Multiple Document QA}''), 
{where the model must retrieve dispersed details based on semantic cues—such as descriptions of shoe style, material, and weight—typically scattered across different parts of the input. These attributes need to be accurately extracted and contextually integrated, challenging the model's ability to maintain coherence and perform fine-grained semantic alignment.
Another example is the comparing product prices task(\ie task ``\texttt{Price}'' ), where the model must identify price points mentioned at different times or in varying contexts, reason about fluctuations, detect potential discounts, and distinguish between original and adjusted prices to support informed comparison.}

\section{Experiments}
\begin{table*}[t]
    \centering
    \small
    \begin{tabular}{c l p{1cm} p{0.6cm}p{0.6cm}p{0.6cm}p{0.6cm}p{0.6cm}p{0.6cm}p{0.6cm}p{0.6cm}p{0.6cm}p{0.8cm}}
    \toprule 
        \multicolumn{12}{c}{\bf \textit{Score}}  \\  
        \cmidrule{3-13}
        &  & \multicolumn{1}{c}{Claimed}  & \multicolumn{3}{c}{Retrieval} & \multicolumn{3}{c}{Hybrid} & \multicolumn{3}{c}{Reasoning} & \multirow{2}{*}{Overall} \\
        &  & Length & Single & Policy & Avg. & Multi & Price & Avg. & Class & Sum. & Avg. \\ \midrule
        & Human & - & 91.5 & 100.0 & 92.1 & 81.8 & 55.4 & 74.9 & 41.0 & 65.8 & 50.0 & 76.3\\ \midrule
        \multirow{4}{*}{\makecell[c]{\rotatebox[origin=c]{0}{\textbf{Closed}}}} 
        & GPT-4o & 128K & 27.7 & 64.0 & 30.2 & 38.9 & \textbf{62.3} & 45.0 & \textbf{87.0} & 75.3 & \textbf{82.7} & 47.6 \\
        & Gemini-1.5-pro & 1000K & \textbf{64.4} & 85.0 & \textbf{65.8} & \textbf{77.7} & 27.9 & \textbf{64.8} & 59.8 & 91.4 & 71.3 & \textbf{66.7} \\
        & Claude-3.7-sonnet & 200K & 47.3 & \textbf{100.0} & 50.8 & 43.3 & 16.5 & 36.3 & 62.5 & 79.3 & 68.6 & 49.9 \\
        & GLM4plus & 131K    & 25.1 & 75.0 & 28.5 & 34.1 & 16.5 & 29.5 & 36.2 & \textbf{92.1} & 56.5 & 35.4\\ 
        \cmidrule(lr){1-13}
        \multirow{4}{*}{\makecell[c]{\rotatebox[origin=c]{0}{\textbf{Open}}}} 
        & Qwen2.5-7B & 131K   & 17.1 & 20.0 & 17.3 & 42.0 & 16.7 & 35.4 & 35.7 & 78.1 & 51.1 & 31.5\\
        & LLaMA-3.1-8B & 128K    & 19.2 & 74.6 & 23.0 & 30.9 & 33.2 & 31.5 & 39.7 & 64.1 & 48.6 & 31.9\\
        & Mistral-7B & 32K  & 9.0  & 80.0 & 13.8 & 33.9 & 13.5 & 28.6 & 33.8 & 52.1 & 40.5 & 25.2\\ 
        & eCeLLM-M & 32K     & 11.5 & 75.0 & 15.8 & 48.4 & 16.9 & 40.2 & 21.4 & 20.0 & 20.9 & 25.6\\ 
        \midrule
        \multicolumn{12}{c}{\bf \textit{Exact Match~(\%)}}  \\  \cmidrule{3-13}
        &  & \multicolumn{1}{c}{Claimed}  & \multicolumn{3}{c}{Retrieval} & \multicolumn{3}{c}{Hybrid} & \multicolumn{3}{c}{Reasoning} & \multirow{2}{*}{Overall} \\
        &  & Length & Single & Policy & Avg. & Multi & Price & Avg. & Class & Sum. & Avg. \\ \midrule
        & Human & - & 89.1 & 100.0 & 89.8 & 51.4 & 15.4 & 42.0 & 4.8 & 8.3 & 6.1 & 53.5 \\ \midrule
        \multirow{4}{*}{\makecell[c]{\rotatebox[origin=c]{0}{\textbf{Closed}}}} 
        & GPT-4o & 128K & 20.6 & 40.0 & 21.9 & \textbf{54.1} & \textbf{61.5} & \textbf{56.0} & 56.0 & 58.6 & \textbf{57.0} & \textbf{42.1} \\
        & Gemini-1.5-pro & 1000K & \textbf{52.3} & 75.0 & \textbf{53.8} & 11.1 & 9.6 & 10.7 & \textbf{59.2} & 44.2 & 53.7 & 38.6 \\
        & Claude-3.7-sonnet & 200K & 34.6 & \textbf{100.0} & 39.0 & 28.4 & 11.8 & 24.0 & 12.5 & \textbf{66.7 }& 32.2 & 32.1 \\
        & GLM4plus & 131K & 18.2 & 75.0 & 22.0 & 21.6 & 7.7 & 18.0 & 0.0 & 50.0 & 18.2 & 19.7\\ 
        \cmidrule(lr){1-13}
        \multirow{4}{*}{\makecell[c]{\rotatebox[origin=c]{0}{\textbf{Open}}}} 
        & Qwen2.5-7B & 131K & 10.9 & 0.0 & 10.2 & 16.2 & 6.7 & 13.7 & 0.0 & 30.8 & 11.2 & 11.7 \\
        & LLaMA-3.1-8B & 128K  & 12.8 & 72.7 & 16.8 & 6.7 & 10.0 & 7.5 & 11.5 & 6.9 & 9.9 & 11.9 \\ 
        & Mistral-7B & 32K & 3.6 & 75.0 & 8.5 & 16.2 & 0.0 & 12.0 & 0.0 & 0.0 & 0.0 & 7.8 \\ 
        & eCeLLM-M & 32K & 5.5 & 75.0 & 10.2 & 35.1 & 7.7 & 28.0 & 0.0 & 0.0 & 0.0 & 14.1\\ 
        \bottomrule
    \end{tabular}
    \caption{Performance comparison of large language models, including closed-source models (GPT-4o (128k), Gemini 1.5 Pro (1000k), Claude 3.7 Sonnet (200k), and GLM4plus (131k)) and widely used open-source models (Qwen, LLaMA, and Mistral). eCeLLM-M is a domain-specific model fine-tuned from Mistral-7B-Instruct.}
    \label{tab:foundation_model}
\end{table*}

Next, we present the evaluation results of LiveLongBench from large language models and context compression methods separately.
Importantly, this section addresses the key questions regarding the capabilities of foundation models and the effectiveness of compression strategies~(as mentioned in Section~\ref{sec-intro}).

\subsection{Large Language Models}

\paragraph{Experimental Setup.}
We investigate whether foundation models can handle long and spoken queries using both closed-source models (GPT-4o, Gemini-1.5-pro, Claude-3.7-sonnet, GLM4plus) and open-source models (Qwen2.5-7B~\footnote{https://huggingface.co/Qwen/Qwen2.5-7B-Instruct}, LLaMA-3.1-8B~\footnote{https://huggingface.co/meta-llama/Llama-3.1-8B-Instruct}, and Mistral-7B~\footnote{https://huggingface.co/mistralai/Mistral-7B-Instruct-v0.3}). Our experimental setup ensures that each model is evaluated under the same conditions, \eg max sequence length. To investigate the impacts of domain-specific fine-tuning, we also include eCeLLM-M~\cite{peng2024ecellm}~\footnote{https://huggingface.co/NingLab/eCeLLM-M}, a model fine-tuned from the large base models Mistral-7B-Instruct for e-commerce, alongside general-purpose LLMs. 
For evaluation metrics, we employ \emph{Exact Match} (\%), which assesses whether the model’s output exactly matches the ground-truth answer. This metric provides a strict evaluation of model correctness. In addition, we introduce a complementary metric, \emph{Score}, which offers a softer and more fine-grained assessment by capturing partial correctness and enabling a continuous measure of model performance across tasks.

\begin{figure}[!h]
\begin{tcolorbox}
\textbf{Findings on Research Question (1)}\\[0.5em]
\emph{While closed-source models remain the strongest, there is a clear gap compared to humans, with retrieval tasks being the most challenging for current models when processing long spoken texts.}
\end{tcolorbox}
\vspace{-1em} 
\end{figure}

\paragraph{Comparison of Foundation Models.}

Table~\ref{tab:foundation_model} presents the performance of various foundation models across tasks. Overall, we find that closed-source models generally outperform open-source ones. For example, Gemini-1.5-Pro achieves the highest overall score (66.7).
However, a notable performance gap remains between LLMs and human annotators, that is even with a 1M-token context window, Gemini-1.5-Pro still falls short of human-level performance (76.3), indicating the benchmark remains challenging.
In addition, we observe that longer context windows substantially improve performance on retrieval-related tasks. Gemini-1.5-Pro significantly outperforms GLM4plus (131k context) in both retrieval (65.8 vs. 28.5) and hybrid tasks (64.8 vs. 29.5), demonstrating the benefit of extended context.
By zooming in open-source models, we observe that Qwen2.5-7B and LLaMA3.1-7B reach overall scores of 31.5 and 31.9, respectively, approaching that of GLM4plus. 
In contrast, other models tend to perform well only on a limited subset of tasks.
Specifically, Qwen2.5-7B performs well on reasoning tasks while LLaMA3.1-7B excels at retrieval.

\begin{figure*}[t]
    \centering
    \includegraphics[width=1.0\linewidth]{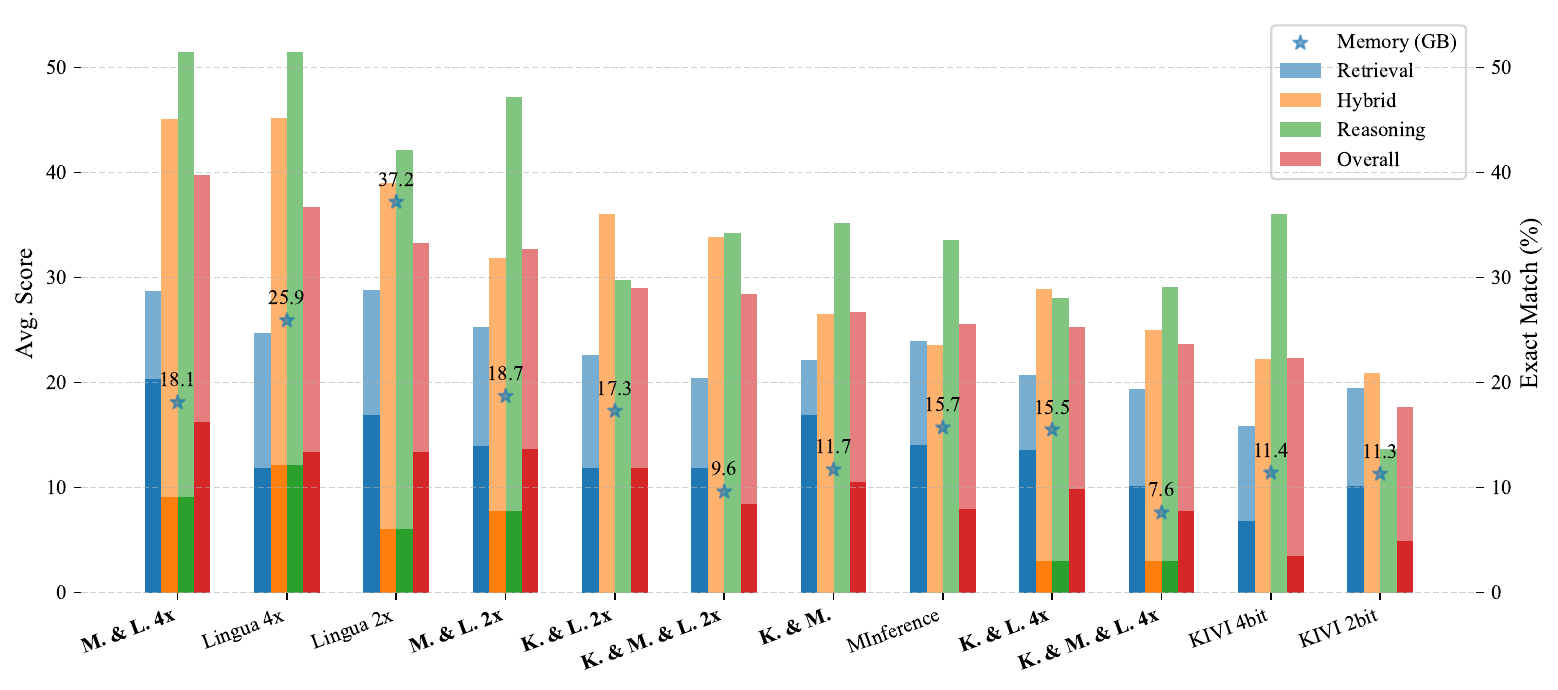}
    \caption{Performance of Context Compression Methods on LLaMA-3.1-8B-Instruct.
  ``K.'' denotes KIVI, ``M.'' denotes MInference, and ``L.'' denotes LLMLingua, while ``2x'' and ``4x'' refer to compression ratios. Methods shown in bold along the x-axis represent multi-methods. From left to right, the methods are arranged in descending order of their Overall average scores. For each bar, the darker segment represents the ``Exact Match (\%)'' score of the corresponding method. Detailed results are provided in Table~\ref{tab:method_main} in the Appendix.}
    \label{fig:Performance}
\end{figure*}

\paragraph{Impacts of Domain-specific Fine-tuning.} 
Models pre-trained or fine-tuned on specialized domains (\eg finance, e-commerce) often exhibit deeper knowledge in those areas, which can enhance reasoning or mitigate redundancy in domain-specific tasks. {As shown in Table~\ref{tab:foundation_model}, although Mistral-7B is not explicitly designed for long-context understanding, domain-specific fine-tuning (eCeLLM-M) still improves its effectiveness on certain long-input tasks. This improvement is likely attributed to its adaptation to domain-specific patterns and structures through fine-tuning.}
Notably, eCeLLM-M demonstrates superior performance in integrated tasks (40.2 in score and 28.0\% exact match), likely due to its enhanced domain understanding. However, this specialization compromises its reasoning ability, resulting in the lowest reasoning score (20.9) and 0.0\% exact match among all evaluated models.

\subsection{Context Compression Methods}
LLMs show varying capabilities in long-context scenarios but often face challenges due to memory usage and computational overhead. To address these limitations, we evaluate existing context compression methods and introduce a simple yet effective baseline for improving their performance.

\paragraph{Experimental Setup.}
We evaluate representative context compression methods on LiveLongBench to assess their utility for long-context understanding and their performance across retrieval, reasoning, and hybrid tasks. The evaluated methods fall into three categories:

$\bullet$~\emph{Token pruning}, which directly removes tokens deemed less relevant, exemplified by LLMLingua~\cite{pan2024llmlingua}.

$\bullet$~\emph{Attention sparsification}, which reduces computational complexity by applying sparse attention mechanisms, represented by MInference~\cite{jiang2024minference}.

$\bullet$~\emph{Quantization}, which compresses internal key-value caches into lower-precision formats, as implemented by KIVI~\cite{liu2024kivi}.

Additionally, we report the performance and resource usage of each model when applying compression methods, ensuring a comprehensive assessment of both accuracy and efficiency. {To illustrate the trade-offs introduced by different strategies, Figure~\ref{fig:Performance} visualizes the performance of LLaMA-3.1-8B-Instruct across multiple compression settings. A detailed summary of the quantitative results is provided in Table~\ref{tab:method_main} in the Appendix.}

\paragraph{Single-Method Analysis.}
Our analysis reveals that different compression methods exhibit distinct preferences across tasks. 1) \emph{Low-bit quantization, by preserving all information, performs better in retrieval tasks, where retaining comprehensive details is critical.} For example, KIVI, even under ultra-low-bit quantization, achieves the highest retrieval accuracy of 80\% in the \texttt{policy} task while maintaining the lowest memory usage. However, its performance declines in other tasks, likely due to excessive compression leading to information loss.
The advantage of KIVI in retrieval is further validated by our experiments on the ``Needle in the Haystack'' task~(See Appendix~\ref{sec-needle}), underscoring the critical role of information retention in achieving accurate retrieval.
2) \emph{In contrast, sparsification and token pruning methods, which discard portions of the input, struggle with retrieval due to incomplete information  
but demonstrate superior reasoning performance.} 
For instance, LLMLingua, with a 4x compression rate, significantly outperforms other single methods in reasoning tasks. This improvement is likely due to the removal of redundant content, which serves as a form of noise reduction, enabling models to focus on essential semantic information.
An empirical case study demonstrates how Lingua4x, by eliminating redundancy, enhances the clarity of key information~(\ie price). 
\begin{figure}[!h]
\begin{tcolorbox}[colback=yellow!10,colframe=gray!85,title=Examples of Denoising Effects of Lingua4x]
\small
$\vartriangleright$ \textbf{Oringinal Text:} \\
\emph{``Let me show you this pair of gloves...''}\\
\textcolor{orange}{<long noisy text>}\\
\emph{``...rabbit wool thermal gloves, just 9.9 yuan per pair! Item No. 1, available for two days. 9.9 yuan per pair, 9.9 yuan per pair!... ''}\\
\textcolor{orange}{<long noisy text>}\\
$\vartriangleright$ \textbf{Compressed Text by Lingua4x:} \\
\emph{``...The Rabbit wool thermal gloves, just 9.9 yuan per pair! 9.9 yuan per pair!...''}\\
\\
$\vartriangleright$ \textbf{Question:} \\
\emph{``What is the price of the rabbit wool thermal gloves?''}\\
$\vartriangleright$ \textbf{w/o Lingua4x Answer:} \\
\textcolor{ErrorRed}{``8.8 yuan''}\\
$\vartriangleright$ \textbf{w Lingua4x Answer:} \\
\textcolor{CorrectGreen}{``The price of the rabbit wool thermal gloves is 9.9 yuan per pair.''}\\
\end{tcolorbox}
\vspace{-0.5em}
\label{example_noise}
\end{figure}

Notably, while compression methods are typically used to reduce computational costs in formal text, our findings reveal that in high-redundancy contexts, they also offer significant denoising effects, improving both model accuracy and overall performance. A case in point is the 4× compression ratio with LLMLingua, which surprisingly outperforms 2× compression (36.7 vs. 33.3), suggesting that higher compression in high-redundancy contexts more effectively suppresses irrelevant information and enhances model performance.

\paragraph{Multi-Methods Analysis.} 
Our analysis highlights that combining different compression strategies can achieve extreme sparsity without compromising performance. As shown in Figure~\ref{fig:Performance}, the \emph{MInference+Lingua4x} combination achieves the highest overall performance by balancing retrieval accuracy and reasoning capabilities. Its strength likely comes from efficient memory utilization and selective token retention. In comparison, \emph{MInference+Lingua2x} excels in reasoning tasks, particularly logical inference, due to its prioritization of critical tokens and attention heads, though with slightly lower retrieval scores. Integrating KIVI with Lingua and MInference maintains competitive retrieval performance but shows weaker reasoning abilities, possibly due to excessive compression affecting long-range coherence.
\begin{figure}[t]
    \centering
    \includegraphics[width=1.0\linewidth]{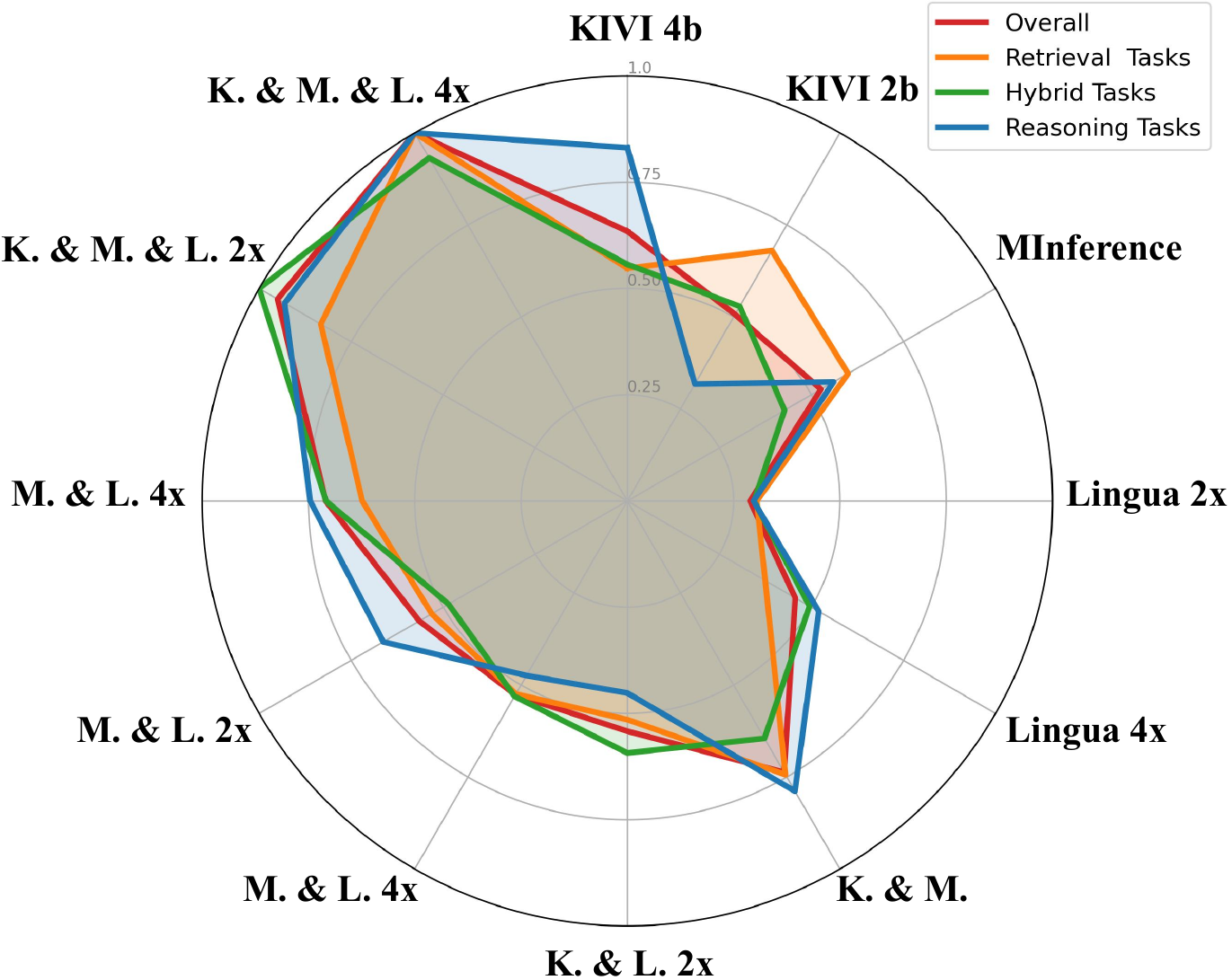}
    \caption{Efficiency Scores Based on DEA Analysis}
    \label{fig:dea}
\end{figure}

\paragraph{Optimal Combination of Balancing Performance and Memory.} 

 As shown in {Figure~\ref{fig:dea}}, to better understand the trade-offs between performance and memory efficiency, we apply \emph{Data Envelopment Analysis (DEA)}, a robust method for evaluating the relative efficiency of different context compression strategies. DEA is a non-parametric approach that treats each method as a Decision-Making Unit (DMU), where memory consumption is considered the input and performance (measured by average score) is the output. By constructing a linear programming model, we assess the efficiency of each compression method, considering both their computational cost and ability to maintain performance across tasks. The resulting efficiency scores, illustrated in the figure, reveal crucial insights: \emph{hybrid approaches, notably the combination of KIVI, Minference, and LLMLingua2x, emerge as the most efficient configuration overall.} This hybrid strategy strikes the best balance, effectively improving performance while minimizing memory usage. The results highlight that hybrid methods outperform individual techniques by integrating complementary strengths, making them an ideal choice for applications like LiveLongBench, where both performance and resource constraints are critical.
\begin{figure}[t]
\begin{tcolorbox}
\textbf{Findings on Research Question (2)}\\[0.5em]
\emph{The combination of Minference and Lingua achieves the best overall performance, while integrating all three methods (KIVI, Minference, and Lingua) strikes the most balanced trade-off between performance and memory efficiency.}
\end{tcolorbox}
\vspace{-1em} 
\end{figure}

\section{Conclusion}
In this work, we introduce LiveLongBench, the first benchmark for evaluating long-context understanding in live-stream spoken texts, featuring sequences of up to 97K tokens. Our evaluation shows that current LLMs suffer notable performance degradation when processing lengthy, informal speech due to redundancy, colloquial expressions, and complex discourse structures. To address these challenges, we found that a hybrid compression strategy that integrates multiple techniques can improve both performance and memory efficiency. Using DEA-based efficiency analysis, we determine the optimal balance among context length, computational cost, and performance. Overall, our study offers new insights into long-context compression and provides practical guidelines for enhancing LLM efficiency in real-world spoken-language applications.

\section{Limitations}
Our work has several limitations. First, LiveLongBench is primarily based on live-streaming content, which may not fully represent the variety of spoken language found in other domains, such as academic lectures or news broadcasts. However, this focus was chosen to capture the dynamic and informal nature of live communication. Second, the evaluation process involves substantial annotation effort, as assessing long-context understanding requires bilingual experts to review extensive documents. Future work should explore automated solutions to reduce this cost while maintaining high evaluation quality.

\bibliography{custom}

\begin{thebibliography}{28}
\providecommand{\natexlab}[1]{#1}

\bibitem[{An et~al.(2023)An, Gong, Zhong, Zhao, Li, Zhang, Kong, and Qiu}]{Leval}
Chenxin An, Shansan Gong, Ming Zhong, Xingjian Zhao, Mukai Li, Jun Zhang, Lingpeng Kong, and Xipeng Qiu. 2023.
\newblock L-eval: Instituting standardized evaluation for long context language models.
\newblock \emph{arXiv preprint arXiv:2307.11088}.

\bibitem[{Bai et~al.(2023)Bai, Lv, Zhang, Lyu, Tang, Huang, Du, Liu, Zeng, Hou et~al.}]{longbench}
Yushi Bai, Xin Lv, Jiajie Zhang, Hongchang Lyu, Jiankai Tang, Zhidian Huang, Zhengxiao Du, Xiao Liu, Aohan Zeng, Lei Hou, et~al. 2023.
\newblock Longbench: A bilingual, multitask benchmark for long context understanding.
\newblock \emph{arXiv preprint arXiv:2308.14508}.

\bibitem[{Chen et~al.(2024)Chen, Zhou, Hua, Loh, Chen, Li, Zhu, and Liang}]{fintextqa}
Jian Chen, Peilin Zhou, Yining Hua, Yingxin Loh, Kehui Chen, Ziyuan Li, Bing Zhu, and Junwei Liang. 2024.
\newblock Fintextqa: A dataset for long-form financial question answering.
\newblock \emph{arXiv preprint arXiv:2405.09980}.

\bibitem[{Coucke et~al.(2018)Coucke, Saade, Ball, Bluche, Caulier, Leroy, Doumouro, Gisselbrecht, Caltagirone, Lavril et~al.}]{snips}
Alice Coucke, Alaa Saade, Adrien Ball, Th{\'e}odore Bluche, Alexandre Caulier, David Leroy, Cl{\'e}ment Doumouro, Thibault Gisselbrecht, Francesco Caltagirone, Thibaut Lavril, et~al. 2018.
\newblock Snips voice platform: an embedded spoken language understanding system for private-by-design voice interfaces.
\newblock \emph{arXiv preprint arXiv:1805.10190}.

\bibitem[{Eisenstein(2013)}]{eisenstein2013what}
Jacob Eisenstein. 2013.
\newblock \href {https://aclanthology.org/N13-1037/} {What to do about bad language on the internet}.
\newblock In \emph{Human Language Technologies: Conference of the North American Chapter of the Association of Computational Linguistics, Proceedings, June 9-14, 2013, Westin Peachtree Plaza Hotel, Atlanta, Georgia, {USA}}, pages 359--369. The Association for Computational Linguistics.

\bibitem[{Fabbri et~al.(2019)Fabbri, Li, She, Li, and Radev}]{fabbri2019multi}
Alexander~R Fabbri, Irene Li, Tianwei She, Suyi Li, and Dragomir~R Radev. 2019.
\newblock Multi-news: A large-scale multi-document summarization dataset and abstractive hierarchical model.
\newblock \emph{arXiv preprint arXiv:1906.01749}.

\bibitem[{Godfrey et~al.(1992)Godfrey, Holliman, and McDaniel}]{godfrey1992switchboard}
John~J Godfrey, Edward~C Holliman, and Jane McDaniel. 1992.
\newblock Switchboard: Telephone speech corpus for research and development.
\newblock In \emph{Acoustics, speech, and signal processing, ieee international conference on}, volume~1, pages 517--520. IEEE Computer Society.

\bibitem[{Hemphill et~al.(1990)Hemphill, Godfrey, and Doddington}]{atis}
Charles~T Hemphill, John~J Godfrey, and George~R Doddington. 1990.
\newblock The atis spoken language systems pilot corpus.
\newblock In \emph{Speech and Natural Language: Proceedings of a Workshop Held at Hidden Valley, Pennsylvania, June 24-27, 1990}.

\bibitem[{Jiang et~al.(2024)Jiang, Li, Zhang, Wu, Luo, Ahn, Han, Abdi, Li, Lin et~al.}]{jiang2024minference}
Huiqiang Jiang, Yucheng Li, Chengruidong Zhang, Qianhui Wu, Xufang Luo, Surin Ahn, Zhenhua Han, Amir~H Abdi, Dongsheng Li, Chin-Yew Lin, et~al. 2024.
\newblock Minference 1.0: Accelerating pre-filling for long-context llms via dynamic sparse attention.
\newblock \emph{arXiv preprint arXiv:2407.02490}.

\bibitem[{Jin et~al.(2024)Jin, Han, Yang, Jiang, Liu, Chang, Chen, and Hu}]{jin2024Selfextend}
Hongye Jin, Xiaotian Han, Jingfeng Yang, Zhimeng Jiang, Zirui Liu, Chia-Yuan Chang, Huiyuan Chen, and Xia Hu. 2024.
\newblock Llm maybe longlm: Self-extend llm context window without tuning.
\newblock \emph{arXiv preprint arXiv:2401.01325}.

\bibitem[{Ko{\v{c}}isk{\`y} et~al.(2018)Ko{\v{c}}isk{\`y}, Schwarz, Blunsom, Dyer, Hermann, Melis, and Grefenstette}]{kovcisky2018narrativeqa}
Tom{\'a}{\v{s}} Ko{\v{c}}isk{\`y}, Jonathan Schwarz, Phil Blunsom, Chris Dyer, Karl~Moritz Hermann, G{\'a}bor Melis, and Edward Grefenstette. 2018.
\newblock The narrativeqa reading comprehension challenge.
\newblock \emph{Transactions of the Association for Computational Linguistics}, 6:317--328.

\bibitem[{Kwan et~al.(2023)Kwan, Zeng, Wang, Sun, Li, Shang, Liu, and Wong}]{m4le}
Wai-Chung Kwan, Xingshan Zeng, Yufei Wang, Yusen Sun, Liangyou Li, Lifeng Shang, Qun Liu, and Kam-Fai Wong. 2023.
\newblock M4le: A multi-ability multi-range multi-task multi-domain long-context evaluation benchmark for large language models.
\newblock \emph{arXiv preprint arXiv:2310.19240}.

\bibitem[{Li et~al.(2017)Li, Su, Shen, Li, Cao, and Niu}]{li2017dailydialog}
Yanran Li, Hui Su, Xiaoyu Shen, Wenjie Li, Ziqiang Cao, and Shuzi Niu. 2017.
\newblock Dailydialog: A manually labelled multi-turn dialogue dataset.
\newblock \emph{arXiv preprint arXiv:1710.03957}.

\bibitem[{Liu et~al.(2024{\natexlab{a}})Liu, Lin, Hewitt, Paranjape, Bevilacqua, Petroni, and Liang}]{liu2024lost}
Nelson~F. Liu, Kevin Lin, John Hewitt, Ashwin Paranjape, Michele Bevilacqua, Fabio Petroni, and Percy Liang. 2024{\natexlab{a}}.
\newblock \href {https://doi.org/10.1162/TACL\_A\_00638} {Lost in the middle: How language models use long contexts}.
\newblock \emph{Trans. Assoc. Comput. Linguistics}, 12:157--173.

\bibitem[{Liu et~al.(2024{\natexlab{b}})Liu, Yuan, Jin, Zhong, Xu, Braverman, Chen, and Hu}]{liu2024kivi}
Zirui Liu, Jiayi Yuan, Hongye Jin, Shaochen Zhong, Zhaozhuo Xu, Vladimir Braverman, Beidi Chen, and Xia Hu. 2024{\natexlab{b}}.
\newblock Kivi: A tuning-free asymmetric 2bit quantization for kv cache.
\newblock \emph{arXiv preprint arXiv:2402.02750}.

\bibitem[{Mohtashami and Jaggi(2023)}]{mohtashami2023passkey}
Amirkeivan Mohtashami and Martin Jaggi. 2023.
\newblock Landmark attention: Random-access infinite context length for transformers.
\newblock \emph{arXiv preprint arXiv:2305.16300}.

\bibitem[{Pan et~al.(2024)Pan, Wu, Jiang, Xia, Luo, Zhang, Lin, R{\"u}hle, Yang, Lin et~al.}]{pan2024llmlingua}
Zhuoshi Pan, Qianhui Wu, Huiqiang Jiang, Menglin Xia, Xufang Luo, Jue Zhang, Qingwei Lin, Victor R{\"u}hle, Yuqing Yang, Chin-Yew Lin, et~al. 2024.
\newblock Llmlingua-2: Data distillation for efficient and faithful task-agnostic prompt compression.
\newblock \emph{arXiv preprint arXiv:2403.12968}.

\bibitem[{Peng et~al.(2024)Peng, Ling, Chen, Sun, and Ning}]{peng2024ecellm}
Bo~Peng, Xinyi Ling, Ziru Chen, Huan Sun, and Xia Ning. 2024.
\newblock ecellm: Generalizing large language models for e-commerce from large-scale, high-quality instruction data.
\newblock \emph{arXiv preprint arXiv:2402.08831}.

\bibitem[{Reid et~al.(2024)Reid, Savinov, Teplyashin, Lepikhin, Lillicrap, Alayrac, Soricut, Lazaridou, Firat, Schrittwieser et~al.}]{reid2024gemini}
Machel Reid, Nikolay Savinov, Denis Teplyashin, Dmitry Lepikhin, Timothy Lillicrap, Jean-baptiste Alayrac, Radu Soricut, Angeliki Lazaridou, Orhan Firat, Julian Schrittwieser, et~al. 2024.
\newblock Gemini 1.5: Unlocking multimodal understanding across millions of tokens of context.
\newblock \emph{arXiv preprint arXiv:2403.05530}.

\bibitem[{Song et~al.(2015)Song, Vallmitjana, Stent, and Jaimes}]{tvsum}
Yale Song, Jordi Vallmitjana, Amanda Stent, and Alejandro Jaimes. 2015.
\newblock Tvsum: Summarizing web videos using titles.
\newblock In \emph{Proceedings of the IEEE conference on computer vision and pattern recognition}, pages 5179--5187.

\bibitem[{Sulem et~al.(2021)Sulem, Hay, and Roth}]{sulem2021we}
Elior Sulem, Jamaal Hay, and Dan Roth. 2021.
\newblock Do we know what we don’t know? studying unanswerable questions beyond squad 2.0.
\newblock In \emph{Findings of the Association for Computational Linguistics: EMNLP 2021}, pages 4543--4548.

\bibitem[{Touvron et~al.(2023)Touvron, Lavril, Izacard, Martinet, Lachaux, Lacroix, Rozi{\`{e}}re, Goyal, Hambro, Azhar, Rodriguez, Joulin, Grave, and Lample}]{touvron2023llama}
Hugo Touvron, Thibaut Lavril, Gautier Izacard, Xavier Martinet, Marie{-}Anne Lachaux, Timoth{\'{e}}e Lacroix, Baptiste Rozi{\`{e}}re, Naman Goyal, Eric Hambro, Faisal Azhar, Aur{\'{e}}lien Rodriguez, Armand Joulin, Edouard Grave, and Guillaume Lample. 2023.
\newblock \href {https://doi.org/10.48550/ARXIV.2302.13971} {Llama: Open and efficient foundation language models}.
\newblock \emph{CoRR}, abs/2302.13971.

\bibitem[{Wang et~al.(2024{\natexlab{a}})Wang, Chen, Cheng, Liao, Zhang, Wu, Yu, Xu, Zhang, Luo, Li, Yang, Huang, and Li}]{wang2024leave}
Minzheng Wang, Longze Chen, Fu~Cheng, Shengyi Liao, Xinghua Zhang, Bingli Wu, Haiyang Yu, Nan Xu, Lei Zhang, Run Luo, Yunshui Li, Min Yang, Fei Huang, and Yongbin Li. 2024{\natexlab{a}}.
\newblock \href {https://aclanthology.org/2024.emnlp-main.322} {Leave no document behind: Benchmarking long-context llms with extended multi-doc {QA}}.
\newblock In \emph{Proceedings of the 2024 Conference on Empirical Methods in Natural Language Processing, {EMNLP} 2024, Miami, FL, USA, November 12-16, 2024}, pages 5627--5646. Association for Computational Linguistics.

\bibitem[{Wang et~al.(2024{\natexlab{b}})Wang, Chen, Cheng, Liao, Zhang, Wu, Yu, Xu, Zhang, Luo et~al.}]{loong}
Minzheng Wang, Longze Chen, Fu~Cheng, Shengyi Liao, Xinghua Zhang, Bingli Wu, Haiyang Yu, Nan Xu, Lei Zhang, Run Luo, et~al. 2024{\natexlab{b}}.
\newblock Leave no document behind: Benchmarking long-context llms with extended multi-doc qa.
\newblock In \emph{Proceedings of the 2024 Conference on Empirical Methods in Natural Language Processing}, pages 5627--5646.

\bibitem[{Williams et~al.(2016)Williams, Raux, and Henderson}]{DSTC}
Jason~D Williams, Antoine Raux, and Matthew Henderson. 2016.
\newblock The dialog state tracking challenge series: A review.
\newblock \emph{Dialogue \& Discourse}, 7(3):4--33.

\bibitem[{Zhang(2018)}]{zhang2018personalizing}
Saizheng Zhang. 2018.
\newblock Personalizing dialogue agents: I have a dog, do you have pets too.
\newblock \emph{arXiv preprint arXiv:1801.07243}.

\bibitem[{Zhang et~al.(2024{\natexlab{a}})Zhang, Chen, Hu, Xu, Chen, Hao, Han, Thai, Wang, Liu et~al.}]{inftybench}
Xinrong Zhang, Yingfa Chen, Shengding Hu, Zihang Xu, Junhao Chen, Moo Hao, Xu~Han, Zhen Thai, Shuo Wang, Zhiyuan Liu, et~al. 2024{\natexlab{a}}.
\newblock $\infty$ bench: Extending long context evaluation beyond 100k tokens.
\newblock In \emph{Proceedings of the 62nd Annual Meeting of the Association for Computational Linguistics (Volume 1: Long Papers)}, pages 15262--15277.

\bibitem[{Zhang et~al.(2024{\natexlab{b}})Zhang, Cao, Ye, Ma, Liao, and Chua}]{TCELongBench}
Zhihan Zhang, Yixin Cao, Chenchen Ye, Yunshan Ma, Lizi Liao, and Tat-Seng Chua. 2024{\natexlab{b}}.
\newblock Analyzing temporal complex events with large language models? a benchmark towards temporal, long context understanding.
\newblock \emph{arXiv preprint arXiv:2406.02472}.

\end{thebibliography}

\appendix

\clearpage
\section{Appendix}
\subsection{Human Annotators.}
To facilitate the evaluation of LLMs, we employed a group of students as human annotators to provide gold-standard labels for the datasets used in our study. These human-generated scores serve as a reference point for comparing the performance of various LLMs. 
Next, we will introduce the annotation process in detail.
\paragraph{Selection of Annotators.}
We selected five students with relevant background knowledge for the task. The annotators have been trained to ensure consistency and accuracy in their labeling, with a focus on the specific requirements of our dataset.
\paragraph{Cost of the Annotation.}
The annotation task was carried out by five full-time students over two days. With each student receiving a monthly salary of 800 RMB, the total cost for this annotation effort amounted to around 400 RMB.
\paragraph{Quality Control.}
To maintain high annotation quality, we conducted regular quality checks throughout the process. This included cross-checking annotations from different annotators and resolving discrepancies through consensus or review by senior researchers.
{Specifically, we applied 6 steps for the quanlity control. 1) Detailed Annotation Guidelines and Training, comprehensive guidelines were developed and training was conducted to ensure clear understanding of the annotation criteria. 2) Pilot Annotation Round, a pilot round was performed on a data subset to refine guidelines and address potential ambiguities.
3) Cross-Checking annotations, each annotation was independently verified by at least two annotators, with discrepancies flagged for review. 4) Consensus-Based resolution, conflicts were resolved through discussion; if consensus could not be reached, senior researchers provided the final decision. 5) Random Sample Review. A random subset of annotations was regularly re-evaluated by independent reviewers to ensure accuracy. 6) Continuous Feedback Loop, regular team meetings provide an ongoing channel for raising concerns and implementing improvements.}
\begin{table}
    \centering
    \small
    \begin{tabular}{lccc}
      \toprule
      \textbf{Task Category} & \textbf{Avg. Token} & \textbf{Lang.} & \textbf{\#Inst.}\\
      \midrule
      \multicolumn{4}{c}{\textit{Task}} \\
      \midrule
      Retrieval & 132.11K & En.\&Zh. & 443 \\
      Reasoning & 20.80K & En.\&Zh. & 129 \\
      Hybrid & 85.07K & En.\&Zh. & 434 \\
      \midrule
      \multicolumn{4}{c}{\textit{Sub Task}} \\
      \midrule
      Single Prod. Retrieval & 147.89K & En.\&Zh. & 351 \\
      Logistics Policy & 71.88K & En.\&Zh. & 92 \\
      Multi Prod. Comparison & 101.47K & En.\&Zh. & 349 \\
      Price Comparison & 17.71K & En.\&Zh. & 85 \\
      Prod. Classification & 20.53K & En.\&Zh. & 21 \\
      Live Summary & 24.38K & En.\&Zh. & 69 \\
      \bottomrule
    \end{tabular}
  \caption{Data statistics of LongLiveBench.}
  \label{tb:stat}
\end{table}
\subsection{Further Analysis of the Data}
\begin{figure*}[t]
    \centering
    \includegraphics[width=1.\linewidth]{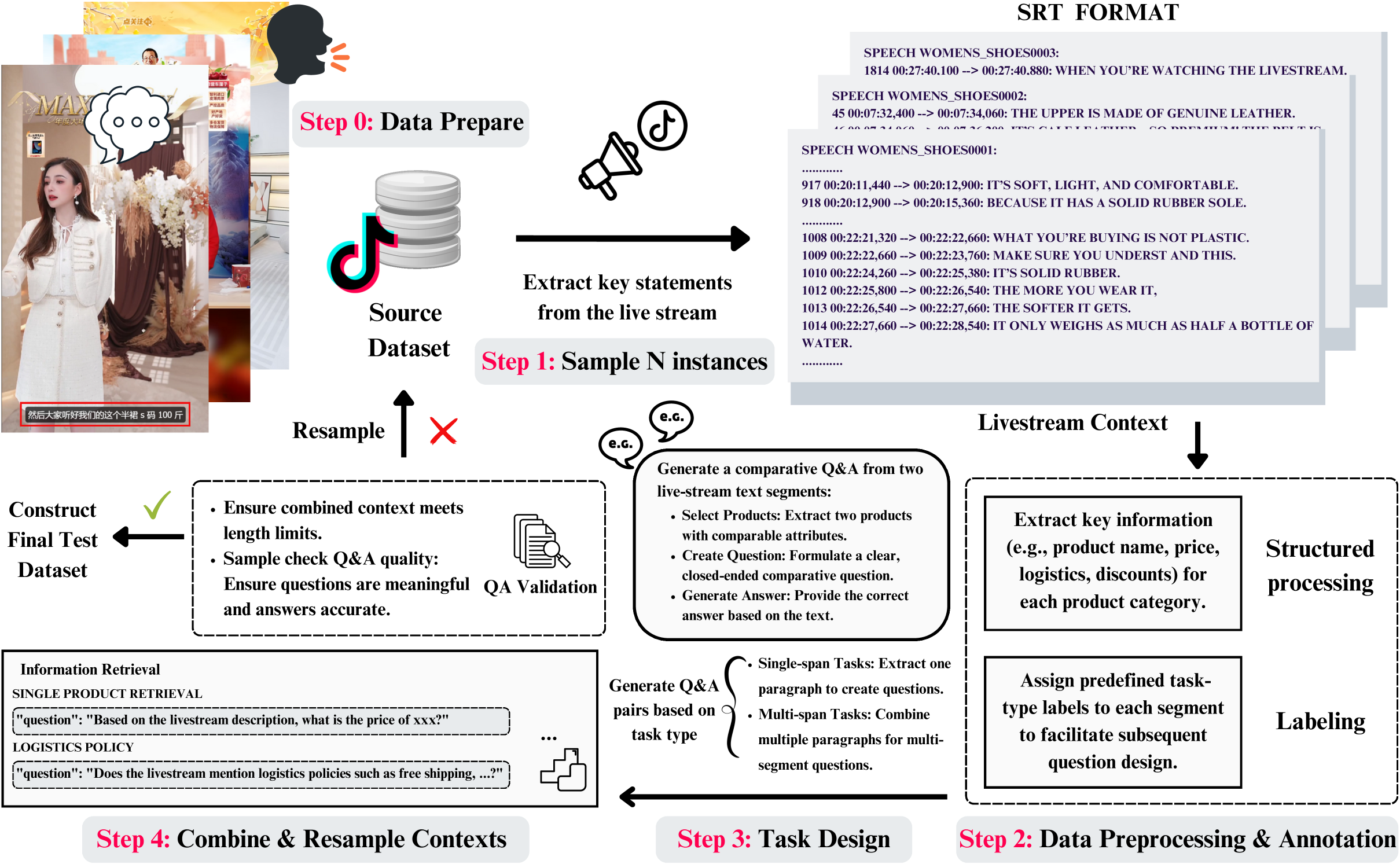}
    \caption{Illustrations of the Construction of LiveLongBench.}
    \label{fig:data_construction}
\end{figure*}

LiveLongBench is constructed through a systematic data collection and processing pipeline, as illustrated in Figure~\ref{fig:data_construction}.  The benchmark integrates multiple task types relevant to long-context understanding in the live-streaming e-commerce domain, ensuring a comprehensive evaluation of large language models.  The detailed statistics of each task within LongLiveBench are presented in Table~\ref{tb:stat}, outlining key dataset characteristics such as the number of instances, average context length, and task-specific attributes.  These details provide a quantitative overview of the dataset composition, highlighting its suitability for assessing KV cache optimization techniques in long-context scenarios.
{\paragraph{ASR performance.}
We evaluated several ASR systems, including Whisper, iFLYTEK, and Paraformer-zh, to obtain accurate results. Using the JiWER package, we compared the transcriptions with human-generated references and calculated the Word Error Rate (WER) and Character Error Rate (CER) for the long live-stream texts (see Figure~\ref{tab:asr}). Although iFLYTEK achieved high accuracy, it supports only 6-hour audio segments. For efficiency, we used Whisper for transcription followed by manual proofreading. After proofreading, Whisper's transcription yielded a WER of 0.53\% and a CER of 0.31\%, confirming the high accuracy of our transcriptions.}
\begin{table}[h]
\centering
\begin{tabular}{lccc}
\toprule
\textbf{ASR} & \textbf{WER} & \textbf{CER} & \textbf{Dur.} \\
\midrule
\textbf{Whisper + Manual}  & \textbf{0.53\%}  & \textbf{0.31\%} &$\infty$ \\
Whisper & 12.74\%  & 10.20\%  &$\infty$  \\
iFLYTEK       & 3.13\%  & 2.35\%  &6h  \\
Paraformer-zh & 36.17\%   &26.69\%  &$\infty$  \\
\bottomrule
\end{tabular}
\caption{Word Error Rate (WER) and Character Error Rate (CER) of different models}
\label{tab:asr}
\end{table}

\begin{figure}[t]
    \centering
    \includegraphics[width=1.0\linewidth]{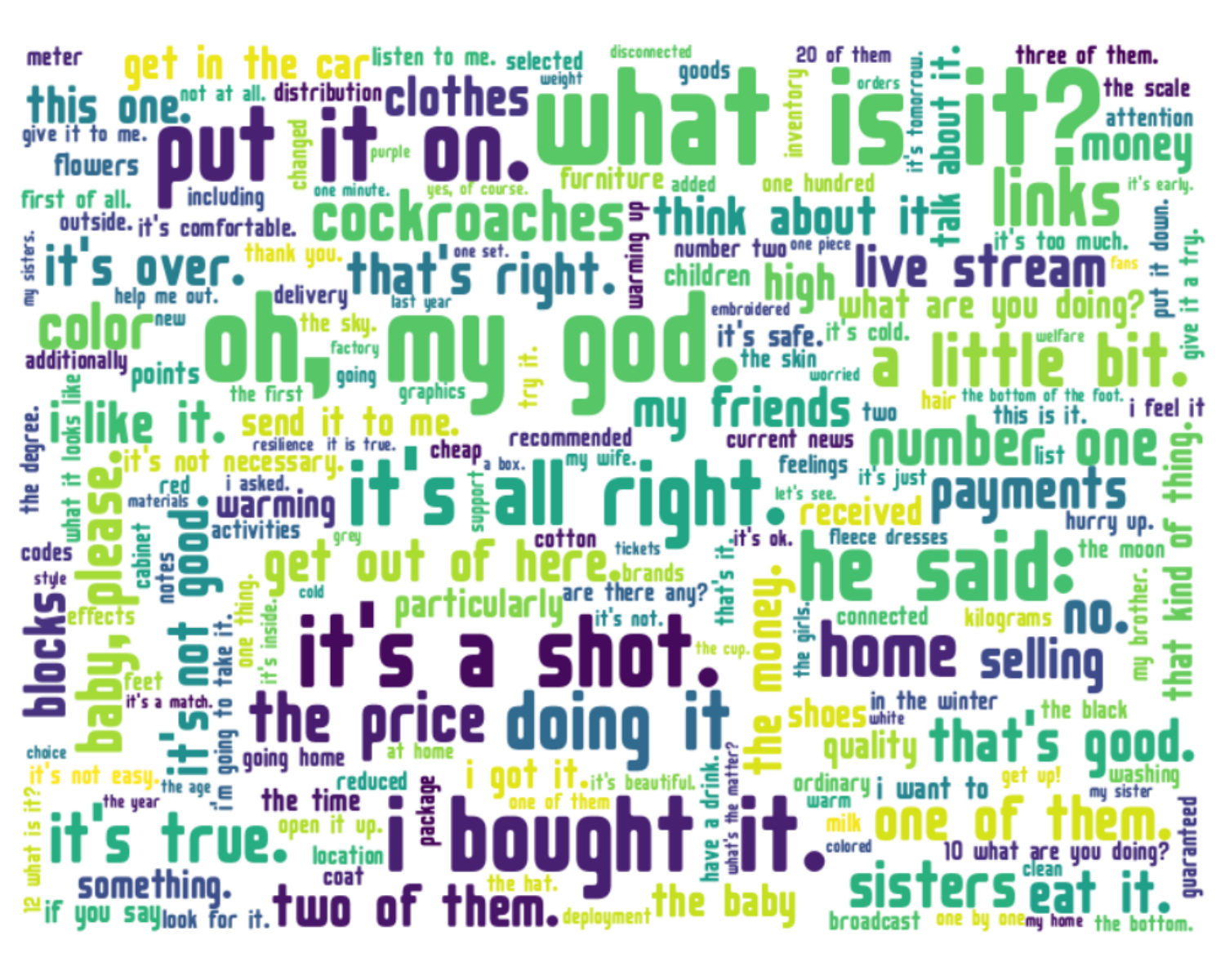}
    \caption{Wordcloud}
    \label{fig:Wordcloud}
\end{figure}

\begin{figure}[t]
    \centering
    \includegraphics[width=0.5\textwidth]{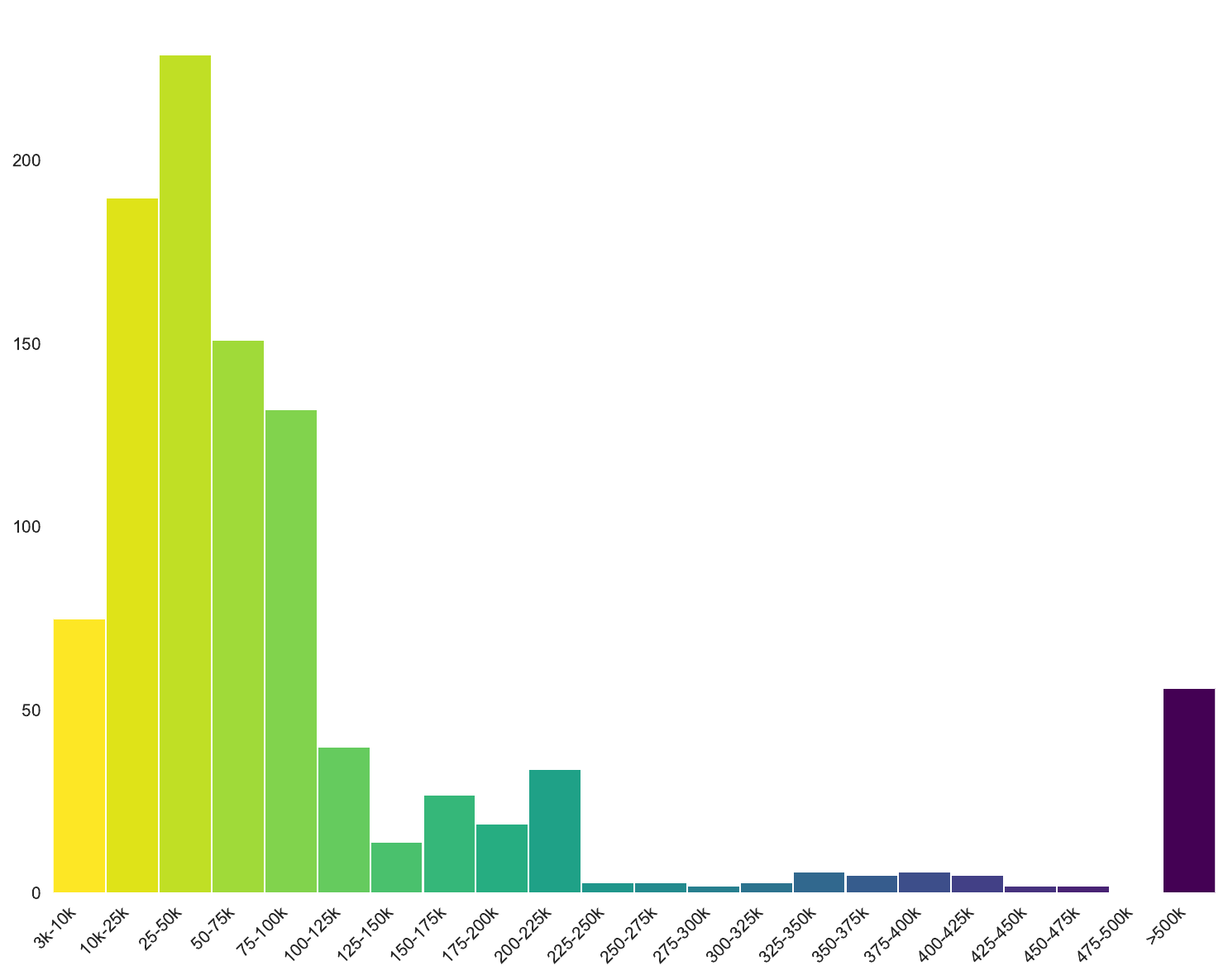}
    \caption{Distributions of the length in LiveLongBench}
    \label{fig:length}
\end{figure}

\paragraph{Length of the Data.}
We present the statistics on the length of LifelongBench. Table~\ref{tb:stat} illustrates the average number of tokens, languages, and test instances across major categories (retrieval, reasoning, hybrid) and their fine-grained subcategories.
In addition, we use a bar plot (see Figure~\ref{fig:length}) to illustrate the distribution of data lengths in LifelongBench. As shown, the data follows a power-law distribution, with the majority of instances concentrated below 220K tokens, while the overall distribution extends beyond 500K tokens.
\paragraph{World Cloud.}
To further explore the dataset, we generate a word cloud representation in Figure~\ref{fig:Wordcloud} that highlights the most frequent terms across the various categories and subcategories of LifelongBench. From this result, we observe a high degree of redundancy in the content, with frequent terms mostly consisting of discourse markers or exclamatory phrases, rather than being closely related to specific content. This observation aligns with the main challenges discussed in Section~\ref{sec-basic}.


\section{Needle-in-a-Haystack Test}
\label{sec-needle}
\begin{figure*}[t]
\setlength{\abovecaptionskip}{0mm}
\setlength{\belowcaptionskip}{0mm}
\centering
\subfigcapskip=-2mm
\subfigure[Baseline]{
    \includegraphics[width=0.23\linewidth]{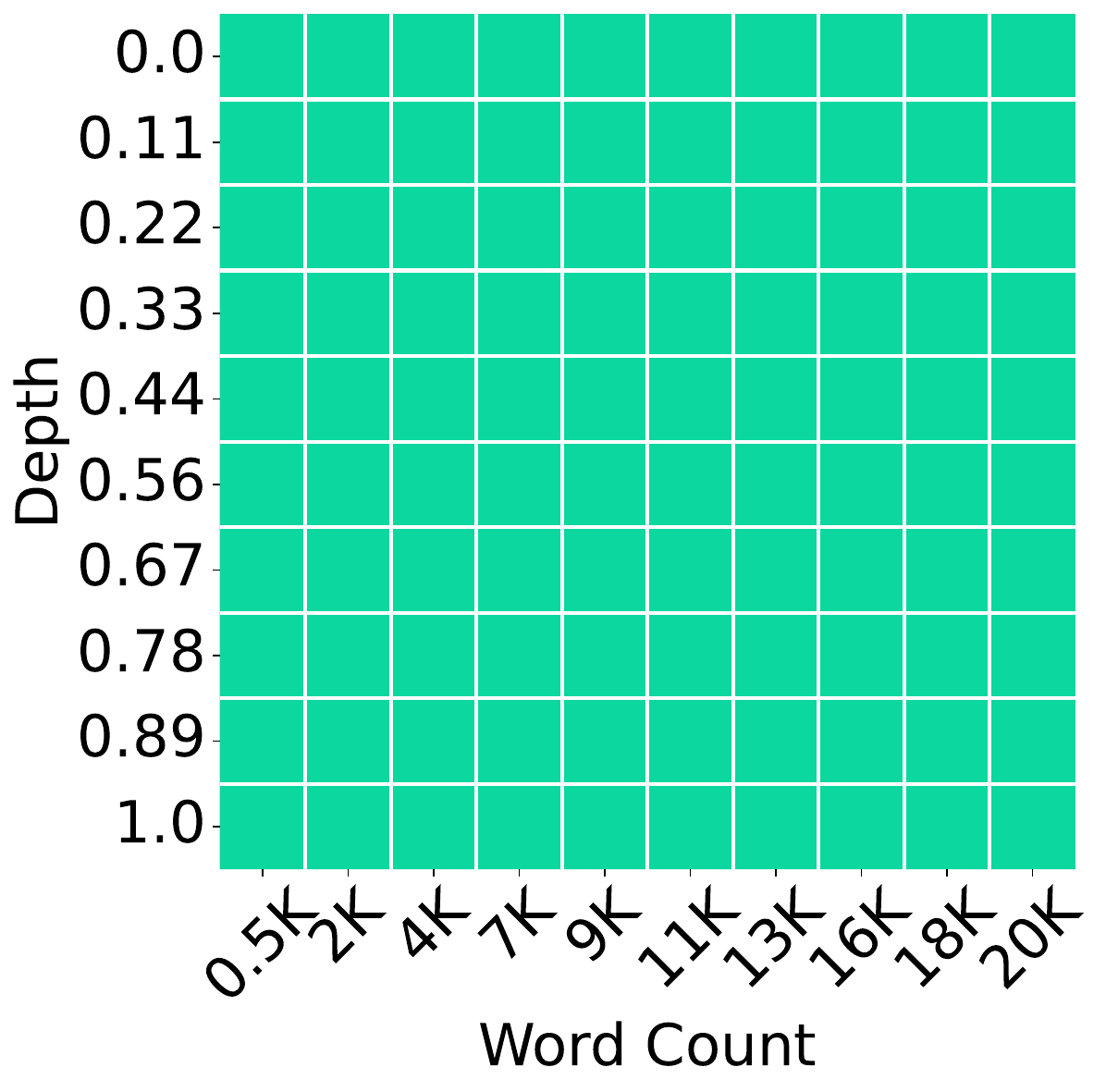}
}
\subfigure[KIVI-2bit]{
    \includegraphics[width=0.23\linewidth]{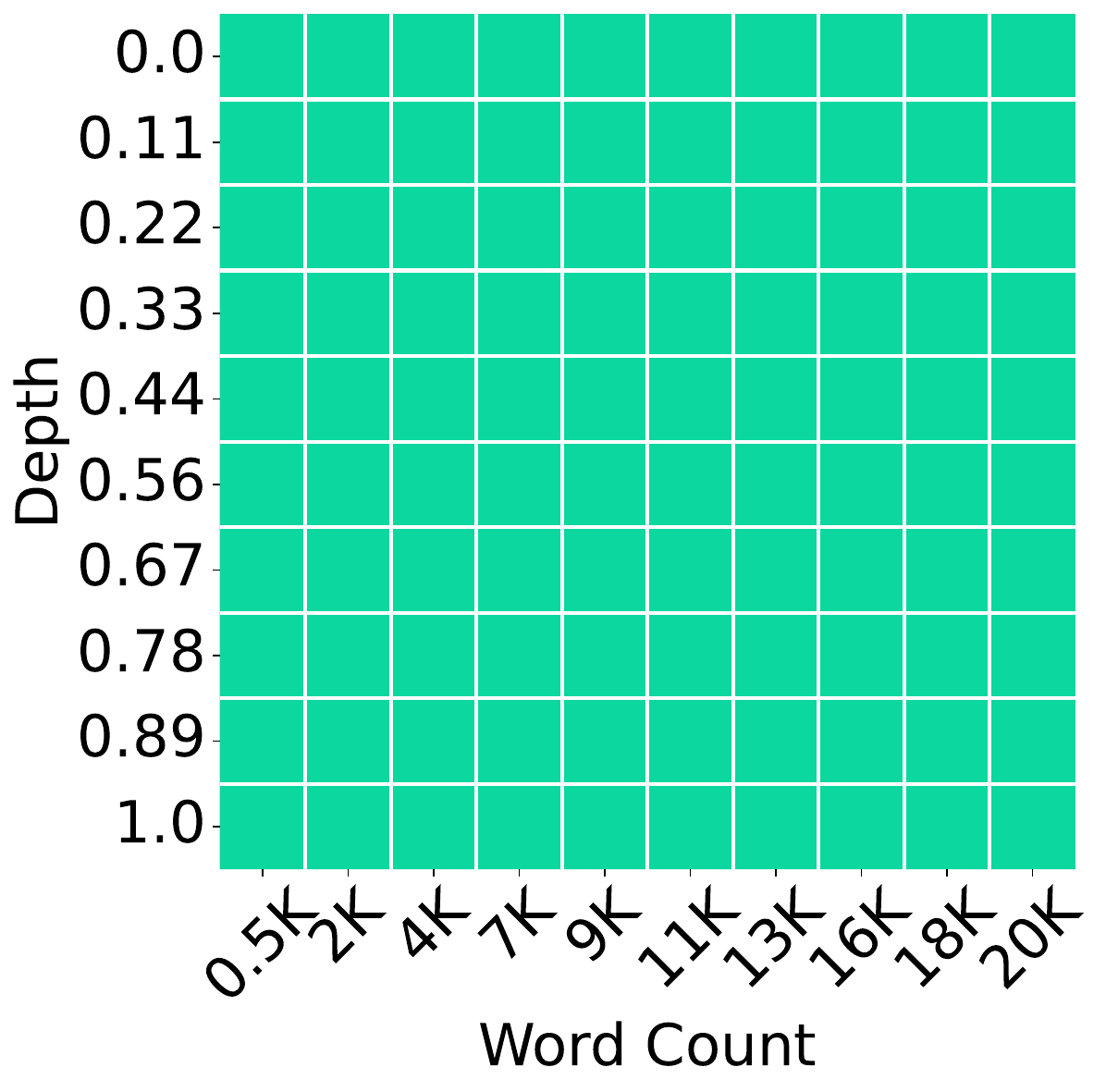}
}
\subfigure[MInference]{
    \includegraphics[width=0.23\linewidth]{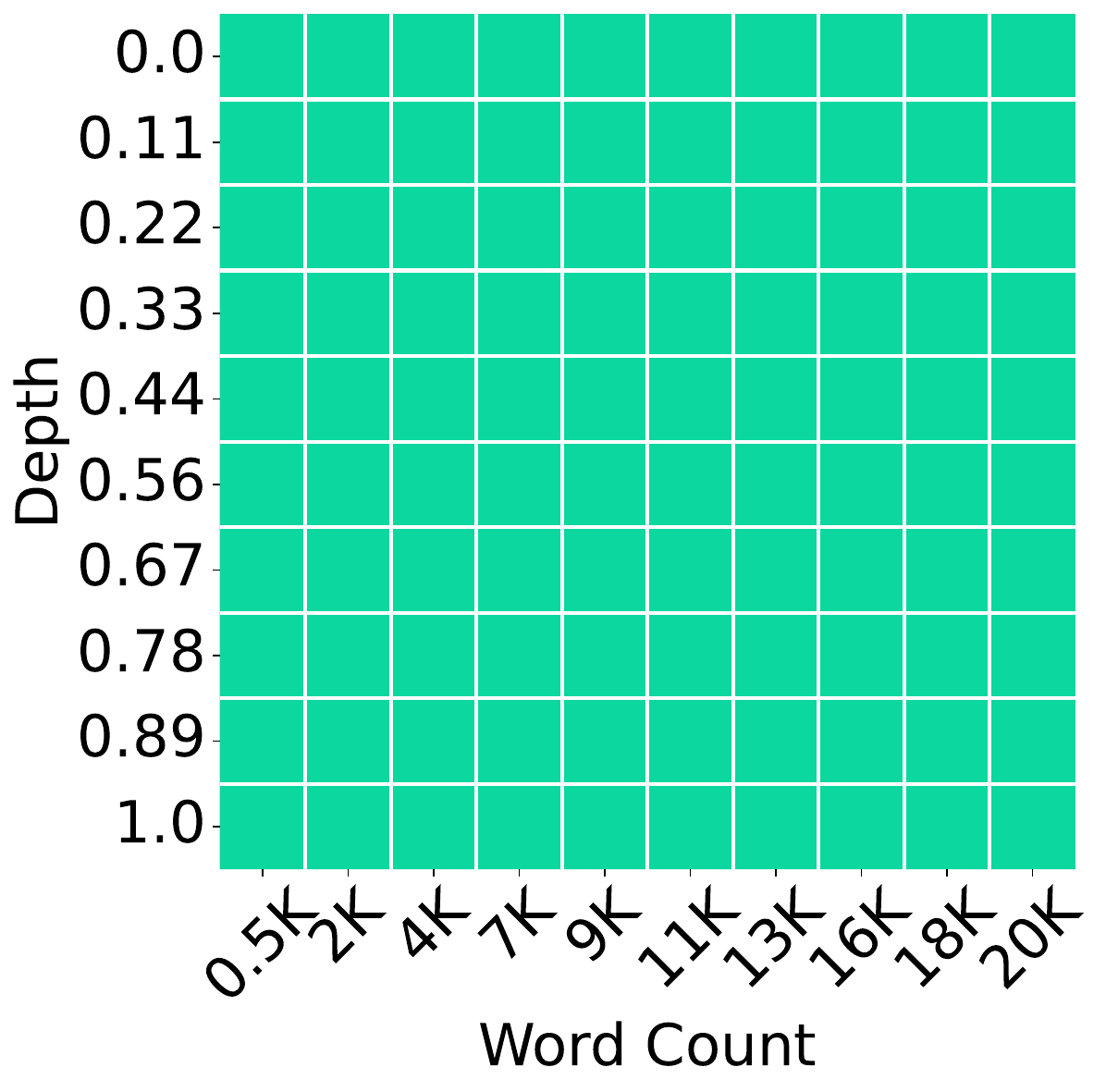}
}
\subfigure[LLMLingua2-2x]{
    \includegraphics[width=0.23\linewidth]{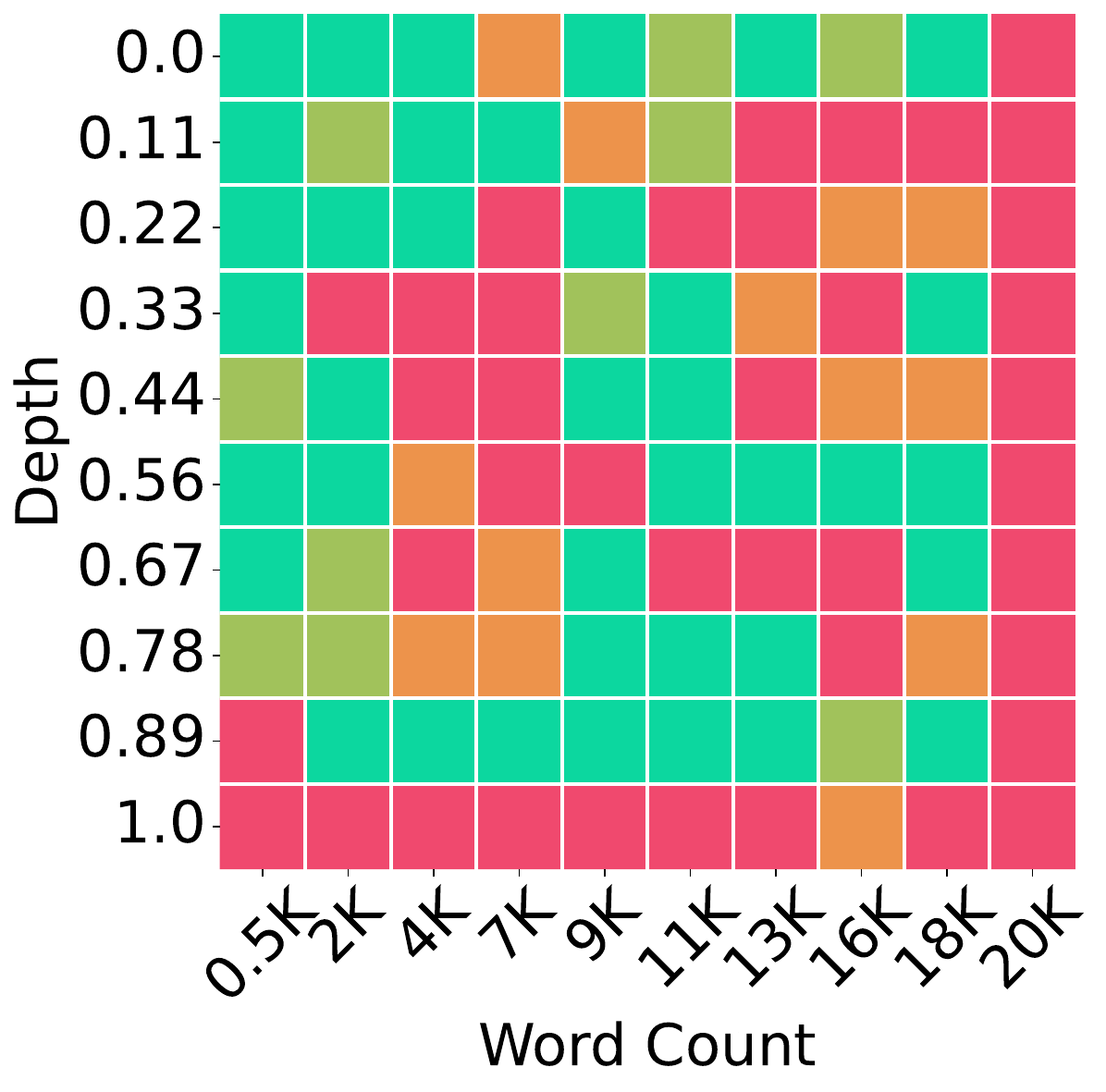}
}
\vspace{-0.5em}
\subfigure[LLMLingua + KIVI]{
    \includegraphics[width=0.23\linewidth]{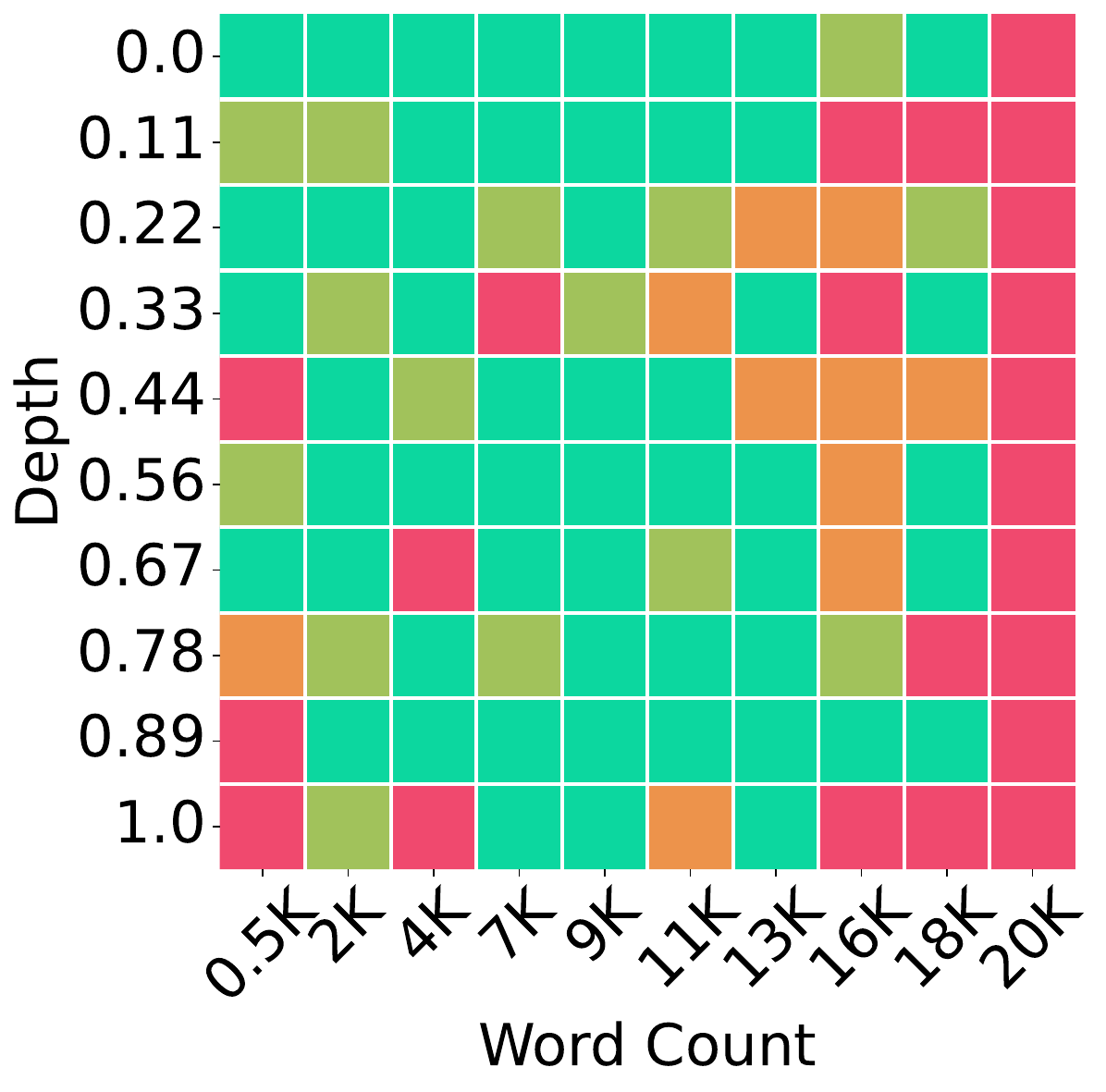}
}
\subfigure[MInference + KIVI]{
    \includegraphics[width=0.23\linewidth]{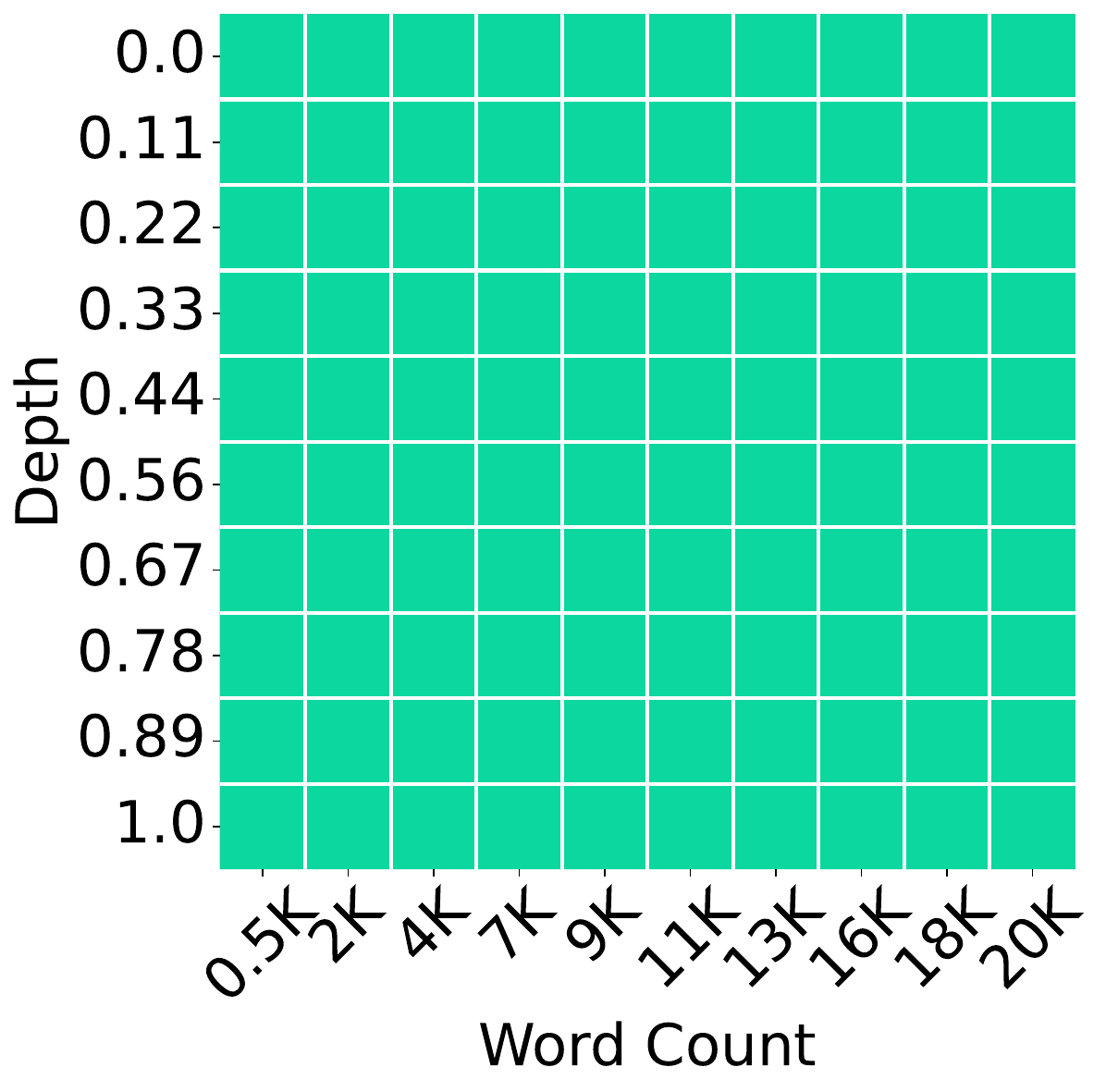}
}
\subfigure[Lingua + MInference]{
    \includegraphics[width=0.23\linewidth]{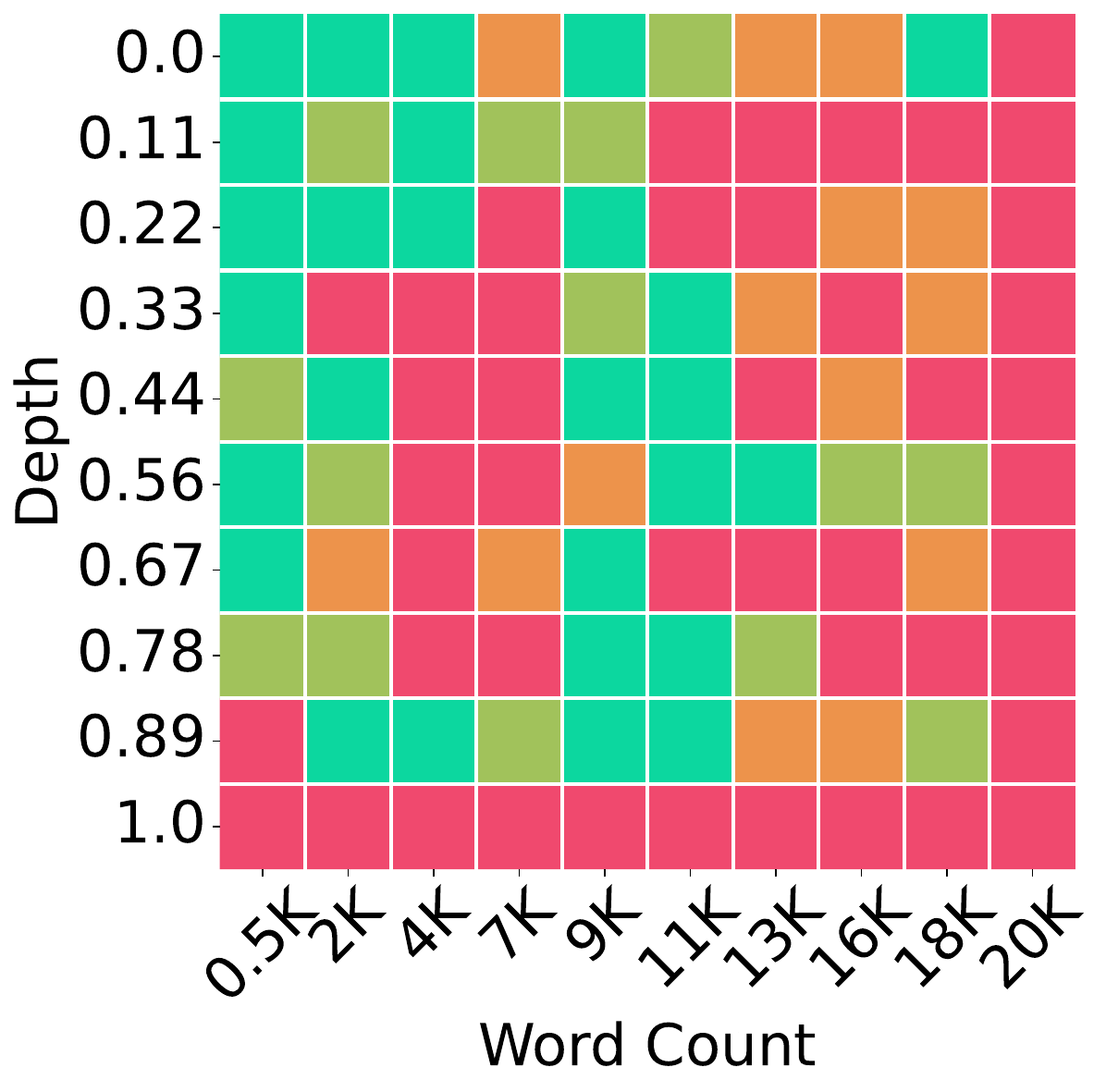}
}
\subfigure[Lingua + MInf + KIVI]{
    \includegraphics[width=0.23\linewidth]{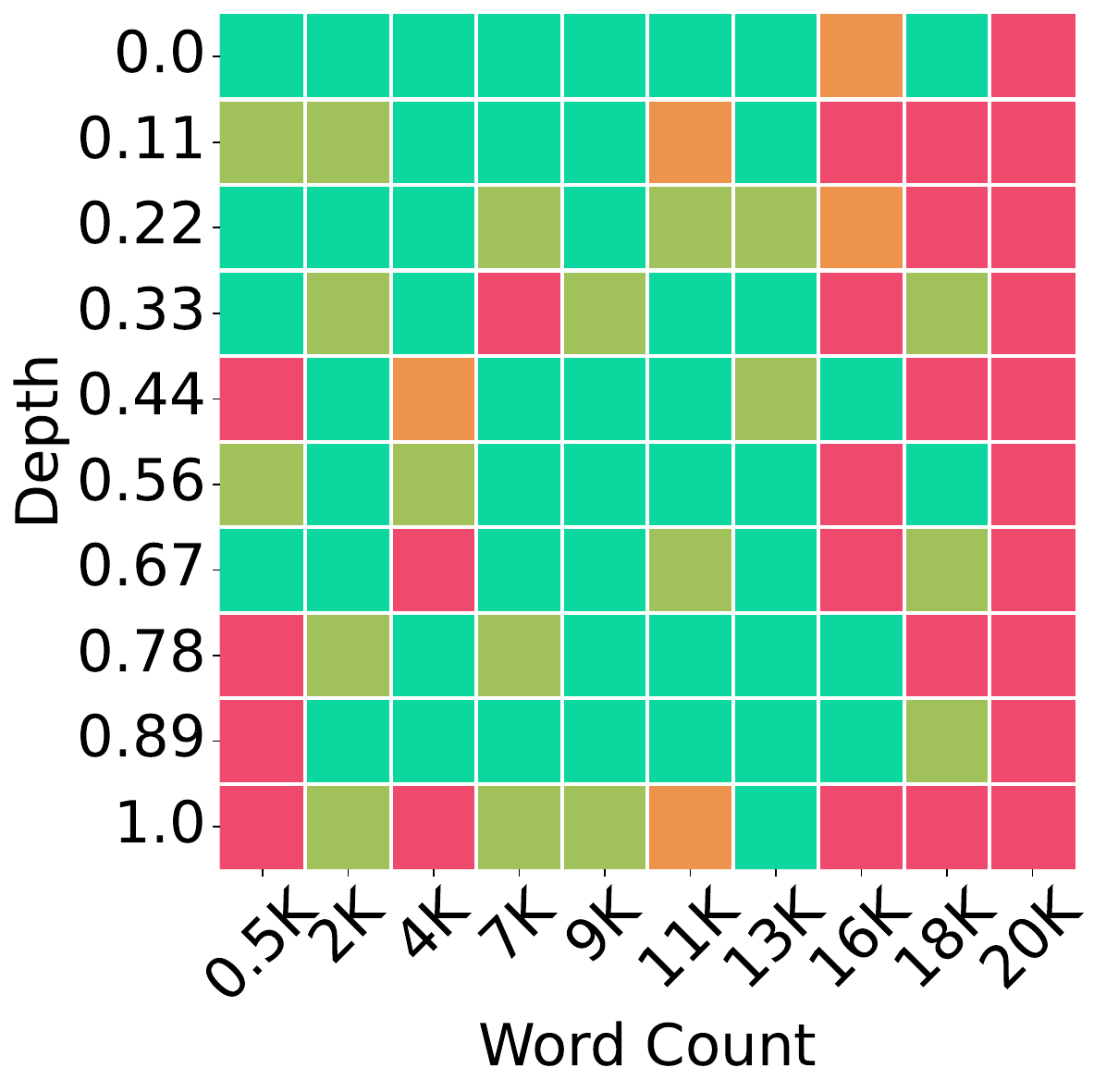}
}
\caption{Needle-in-a-Haystack results for each method on Llama-3-8B-Instruct. where a 20k words length input is converted to approximately 28k tokens.}
\label{fig:needle_main}
\end{figure*}

\paragraph{Experimental Setup.}
We follow the work~\cite{mohtashami2023passkey} to execute the Needle-in-a-Haystack Test. The corpus comprises live stream transcripts characterized by high redundancy and informal, spoken language. The results are presented in the Figure~\ref{fig:needle_main}.

\paragraph{Results.}
Our results highlight the unique advantage of low-bit quantization in preserving retrieval performance, aligning with previous findings that retaining more information is critical for accurate retrieval. KIVI effectively reduces memory usage while maintaining retrieval accuracy, reinforcing the importance of information retention in long-context tasks.
In addition, we also observe that the combination of MInference+KIVI consistently achieves strong retrieval performance, validating the effectiveness of hybrid compression methods in balancing efficiency and accuracy.

\subsection{Needle-in-a-Haystack Test Details}
Needle-in-a-Haystack (NIAH) a style of synthetically generated stress test designed to assess a language model’s ability to retrieve specific information embedded within a large volume of unrelated background text. The core task involves inserting a critical piece of information at varying positions within different lengths of irrelevant content and then querying the model to recall this information accurately. Specifically,\citet{mohtashami2023passkey} introduced a standardized passkey retrieval task, in which a key phrase formatted as “The pass key is <PASS KEY>. Remember it. <PASS KEY> is the pass key” is inserted into background text composed of repetitive generic sentences such as “The grass is green. The sky is blue. The sun is yellow. Here we go. There and back again.” This formulation ensures that the task is purely focused on retrieval rather than inference. A variation of NIAH proposed by  \href{https://github.com/gkamradt/LLMTest_NeedleInAHaystack}{Greg Kamradt} replaces the passkey with a more natural sentence, such as “The best thing to do in San Francisco is eat a switch and sit in Dolores Park on a sunny day,” which serves as the retrievable target. In both formulations, the objective for large language models (LLMs) remains the same: they must successfully extract the inserted key information from an overwhelming amount of distractor text. Our implementation of the NIAH task closely follows the passkey retrieval template proposed by \citet{mohtashami2023passkey}. However, we introduce two key modifications: (1) the use of a 7-digit passkey instead of a generic phrase, and (2) the replacement of artificially structured background text with colloquial multi-domain live-streaming transcript fragments. This adjustment more closely reflects real-world applications where models must filter out irrelevant conversational noise while preserving and retrieving critical embedded information. As described in \href{https://github.com/gkamradt/LLMTest_NeedleInAHaystack}{Arize-ai} and \citet{reid2024gemini}, the general retrieval prompt structure follows:
"There is an important piece of information hidden inside a large volume of irrelevant text. Your task is to find and memorize it. I will later quiz you about this information."
A standard filler, such as excerpts from Paul Graham’s essays, precedes the inserted passkey phrase:
\texttt{"The pass key is <7-DIGIT PASS KEY>. Remember it. <7-DIGIT PASS KEY> is the pass key."
A suffix filler follows, after which the model is prompted with:
"What is the pass key?"}

\section{Optimal Combination of Compression Methods with the Effect of SelfExtend}
To evaluate the effectiveness of various KV cache compression methods and their combinations, we conduct experiments on LiveLongBench using LLaMA-3.1-8B-Instruct.  The results, presented in Table~\ref{tab:method_main}, illustrate the performance of individual compression techniques as well as hybrid approaches, providing insights into their impact on long-context processing.  The table details the accuracy and overall scores achieved under different configurations, highlighting the trade-offs between compression efficiency and model performance.  
\begin{table*}[t]
    \centering
    \small
    \begin{tabular}{lllp{0.6cm}p{0.6cm}p{0.6cm}p{0.6cm}p{0.6cm}p{0.6cm}p{0.6cm}p{0.6cm}p{0.6cm}p{0.6cm}p{0.8cm}}
    \toprule 
        & & \multicolumn{11}{c}{\bf \textit{Score}}  \\  \cmidrule{3-12}
        & Mem.& \multicolumn{3}{c}{Retrieval} & \multicolumn{3}{c}{Hybrid} & \multicolumn{3}{c}{Reasoning} & \multirow{2}{*}{Overall}\\
        & (GB) & Single & Policy  & Avg. & Multi & Price & Avg. & Class  & Sum. & Avg. \\ \midrule
        Full & OOM & -  & - & - & - & - & - & - & - & - & - \\ \midrule
        \ding{172}~KIVI 4bit & 11.4 & 11.2 & \textbf{80.0} & 15.8 & 22.7 & 16.2 & 21.0 & 23.1 & 58.8 & 36.1 & 22.4 \\
        \ding{173}~KIVI 2bit & 11.2 & 15.1 & \textbf{80.0} & 19.5 & 21.9 & 7.7 & 18.2 & 10.5 & 19.2 & 13.6 & 17.7 \\
        \ding{174}~Minference & 15.7 & 21.6 & 57.1 & 24.0 & 24.1 & 17.0 & 22.2 & 19.5 & 58.3 & 33.6 & 25.6 \\
        \ding{175}~Lingua 2x & 37.2 & \textbf{26.0} & 67.5 & \textbf{28.8} & 41.2 & 8.5 & 32.7 & 28.3 & 66.3 & 42.1 & 33.3 \\
        \ding{176}~Lingua 4x & 25.9 & 22.7 & 52.5 & 24.7 & \textbf{46.6} & \textbf{25.0} & \textbf{41.0} & \textbf{39.5} & \textbf{72.5} & \textbf{51.5} & \textbf{36.7} \\ \midrule
        \ding{172}+\ding{174} & 11.7 & 18.3 & \textbf{75.0} & 22.1 & 29.3 & 18.5 & 26.5 & 22.9 & 56.7 & 35.2 & 26.7 \\
        \ding{172}+\ding{175} & 17.3 & 19.9 & 60.0 & 22.6 & 36.1 & 35.9 & 36.0 & 15.4 & 55.0 & 29.8 & 29.0 \\
        \ding{172}+\ding{176} & 15.5 & 17.8 & 60.0 & 20.7 & 35.0 & 11.5 & 28.9 & 12.1 & 55.8 & 28.0 & 24.7 \\
        \ding{174}+\ding{175} & 18.7 & 22.6 & 61.7 & 25.2 & 34.6 & 24.1 & 31.9 & 28.1 & 80.7 & 47.2 & 32.7 \\
        \ding{174}+\ding{176} & 18.1 & \textbf{26.4} & 61.3 & \textbf{28.7} & \textbf{46.1} & \textbf{42.3} & \textbf{45.1} & \textbf{34.5} & \textbf{81.3} & \textbf{51.5} & \textbf{39.8} \\
        \midrule
        \ding{172}+\ding{174}+\ding{175} & 9.6 & 17.6 & \textbf{60.0} & \textbf{20.4} & \textbf{34.7} & \textbf{31.5} & \textbf{33.9} & \textbf{18.6} & \textbf{61.7} & \textbf{34.2} & \textbf{28.4} \\
        \ding{172}+\ding{174}+\ding{176} & 7.6 & \textbf{17.9} & 40.0 & 19.4 & 28.6 & 14.6 & 25.0 & 12.1 & 58.8 & 29.1 & 23.6 \\
        \midrule
        & & \multicolumn{11}{c}{\bf \textit{Exact Match~(\%)}}  \\  \cmidrule{3-12}
        & Mem. & \multicolumn{3}{c}{Retrieval} & \multicolumn{3}{c}{Hybrid} & \multicolumn{3}{c}{Reasoning} & \multirow{2}{*}{Overall}\\
        & (GB) & Single & Policy  & Avg. & Multi & Price & Avg. & Class  & Sum. & Avg. \\ \midrule
        Full & OOM & -  & - & - & - & - & - & - & - & - & - \\ \midrule
        \ding{172}~KIVI 4bit & 11.4 & 1.8 & 75.0 & 6.8 & 2.7 & 0.0 & 2.0 & 0.0 & 0.0 & 0.0 & 3.5 \\
        \ding{173}~KIVI 2bit & 11.2 & 5.5 & 75.0 & 10.2 & 2.7 & 0.0 & 2.0 & 0.0 & 0.0 & 0.0  & 4.9 \\
        \ding{174}~Minference & 15.7 & 10.9 & 57.1 & 14.0 & 8.1 & 0.0 & 6.0 & 0.0 & 0.0 & 0.0  & 8.0 \\
        \ding{175}~Lingua 2x & 37.2 & 14.6 & 50.0 & 17.0 & 18.9 & 0.0 & 14.0 & 0.0 & 16.7 & 6.1  & 13.4 \\
        \ding{176}~Lingua 4x & 25.9 & 10.9 & 25.0 & 11.9 & 18.9 & 7.7 & 16.0 & 14.3 & 8.3 & 12.1  & 13.4 \\
        \midrule
        \ding{172}+\ding{174} & 11.7 & 12.7 & 75.0 & 17.0 & 10.8 & 7.7 & 10.0 & 0.0 & 0.0 & 0.0 & 10.6 \\
        \ding{172}+\ding{175} & 17.3 & 9.1 & 50.0 & 11.9 & 16.2 & 29.4 & 19.7 & 0.0 & 0.0 & 0.0 & 11.9 \\
        \ding{172}+\ding{176} & 15.5 & 10.9 & 50.0 & 13.6 & 13.5 & 0.0 & 10.0 & 0.0 & 8.3 & 3.0 & 9.9 \\
        \ding{174}+\ding{175} & 18.7 & 12.5 & 33.3 & 13.9 & 20.0 & 9.5 & 17.3 & 4.8 & 13.0 & 7.8 & 13.7 \\
        \ding{174}+\ding{176} & 18.1 & 20.0 & 25.0 & 20.3 & 16.2 & 15.4 & 16.0 & 9.5 & 8.3 & 9.1 & 16.2 \\
        \midrule
        \ding{172}+\ding{174}+\ding{175} & 9.6 & 9.1 & 50.0 & 11.9 & 10.8 & 7.7 & 10.0 & 0.0 & 0.0 & 0.0 & 8.5 \\
        \ding{172}+\ding{174}+\ding{176} & 7.6 & 9.1 & 25.0 & 10.2 & 10.8 & 0.0 & 8.0 & 12.5 & 8.3 & 4.9 & 7.8 \\
        \bottomrule
    \end{tabular}
    
    \caption{Performance of context compression methods on LLaMA-3.1-8B-Instruct.}
    \label{tab:method_main}
\end{table*}

Building upon our evaluation of KV cache compression methods, we further explore the integration of Self-Extend~\cite{jin2024Selfextend}, a self-regressive extension technique designed to enhance inference by expanding the context window of existing LLMs. As shown in Table~\ref{tab:method+selfE}, we incorporate Self-Extend into two compression method combinations: (1) the performance-optimal configuration, “MInference (\ding{174}) + LLMLingua 4× (\ding{176})”, and (2) the resource-performance balanced configuration, “KIVI 4-bit (\ding{172}) + MInference (\ding{174}) + LLMLingua 4× (\ding{176})”, identified using the DEA method. In the table, different compression methods are denoted as follows: \ding{172} for KIVI, \ding{174} for MInference, \ding{175} for LLMLingua 2×, \ding{176} for LLMLingua 4×, and \ding{177} for Self-Extend. Experimental results demonstrate that incorporating Self-Extend (\ding{177}) into the resource-optimal method further enhances inference performance, reinforcing the model’s ability to process long-context inputs effectively.

\begin{table*}[t]
    \centering
    \small
    \begin{tabular}{lllp{0.6cm}p{0.6cm}p{0.6cm}p{0.6cm}p{0.6cm}p{0.6cm}p{0.6cm}p{0.6cm}p{0.6cm}p{0.8cm}}
    \toprule 
        & & \multicolumn{11}{c}{\bf \textit{Score}}  \\  \cmidrule{3-12}
        & Mem.& \multicolumn{3}{c}{Retrieval} & \multicolumn{3}{c}{Hybrid} & \multicolumn{3}{c}{Reasoning} & \multirow{2}{*}{Overall}\\
        & (GB) & Single & Policy  & Avg. & Multi & Price & Avg. & Class  & Sum. & Avg. \\ \midrule
        \ding{174}+\ding{176} & 18.1 & \textbf{26.4} & 61.3 & \textbf{28.7} & \textbf{46.1} & 42.3 & \textbf{45.1} & 34.5 & \textbf{81.3} & \textbf{51.5} & \textbf{39.8} \\ 
        \ding{174}+\ding{176}+\ding{177} & 18.1 & 18.7 & \textbf{68.8} & 22.1 & 39.7 & \textbf{43.5} & 40.7  & \textbf{36.7} & 72.1 & 50.0 & 35.0 \\ \midrule
        \ding{172}+\ding{174}+\ding{176} & 7.6 & \textbf{17.9} & 40.0 & \textbf{19.4} & 28.6 & 14.6 & 25.0 & 12.1 & 58.8 & 29.1 & 23.6 \\
        \ding{172}+\ding{174}+\ding{176}+\ding{177} & 7.6 & 14.6 & \textbf{52.5} & 17.2 & \textbf{29.7} & \textbf{21.5} & \textbf{27.6} & \textbf{21.4} & \textbf{60.8} & \textbf{35.8} & \textbf{25.2} \\
        \midrule

        & & \multicolumn{11}{c}{\bf \textit{Exact Match~(\%)}}  \\  \cmidrule{3-12}
        & Mem. & \multicolumn{3}{c}{Retrieval} & \multicolumn{3}{c}{Hybrid} & \multicolumn{3}{c}{Reasoning} & \multirow{2}{*}{Overall}\\
        & (GB) & Single & Policy  & Avg. & Multi & Price & Avg. & Class  & Sum. & Avg. \\ \midrule
        \ding{174}+\ding{176} & 18.1 & 20.0 & 25.0 & 20.3 & 16.2 & 15.4 & 16.0 & 9.5 & 8.3 & 9.1 & 16.2 \\
        \ding{174}+\ding{176}+\ding{177} & 18.1 & 12.7 & 25.0 & 13.6 & 16.2 & 7.7 & 14.0  & 14.3 & 25.0 & 18.2 & 14.8 \\ \midrule
        \ding{172}+\ding{174}+\ding{176} & 7.6 & 9.1 & 25.0 & 10.2 & 10.8 & 0.0 & 8.0 & 12.5 & 8.3 & 4.9 & 7.8 \\
        \ding{172}+\ding{174}+\ding{176}+\ding{177} & 7.6 & 3.6 & 25.0 & 5.1 & 10.8 & 7.7 & 10.0 & 9.5 & 9.1 & 7.3 & 7.8 \\
        \bottomrule
    \end{tabular}
    
    \caption{Optimal Combination of Compression Methods with the Effect of SelfExtend}
    \label{tab:method+selfE}
\end{table*}

\section{Case Study on the Performance of Different Compression Methods.}
\begin{figure*}[t]
    \centering
    \includegraphics[width=1.0\linewidth]{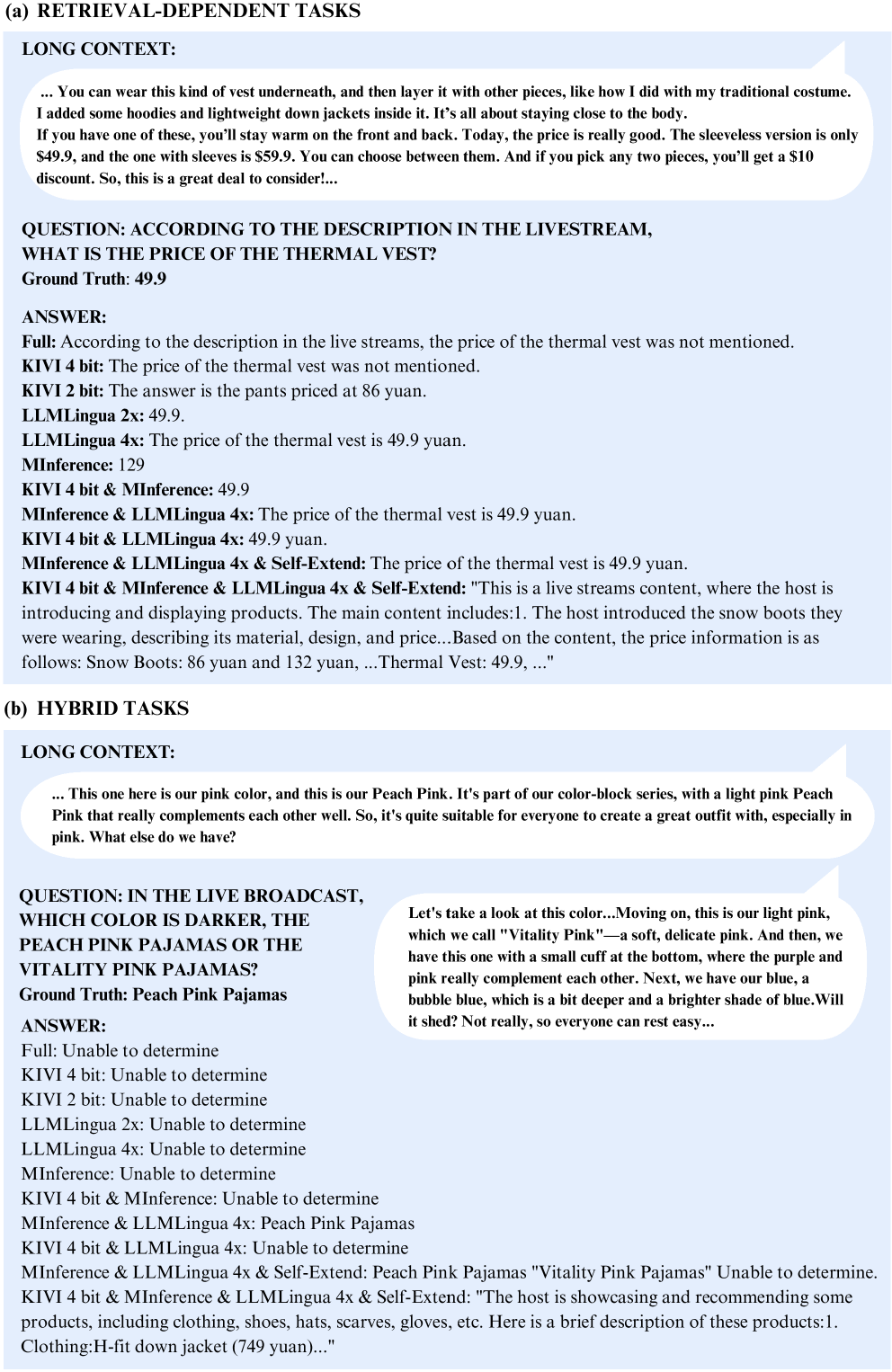}
    \caption{Case Study of Retrieval-Dependent \& Hybrid Tasks.}
    \label{fig:method_case_1}
\end{figure*}

\begin{figure*}[t]
    \centering
    \includegraphics[width=1.0\linewidth]{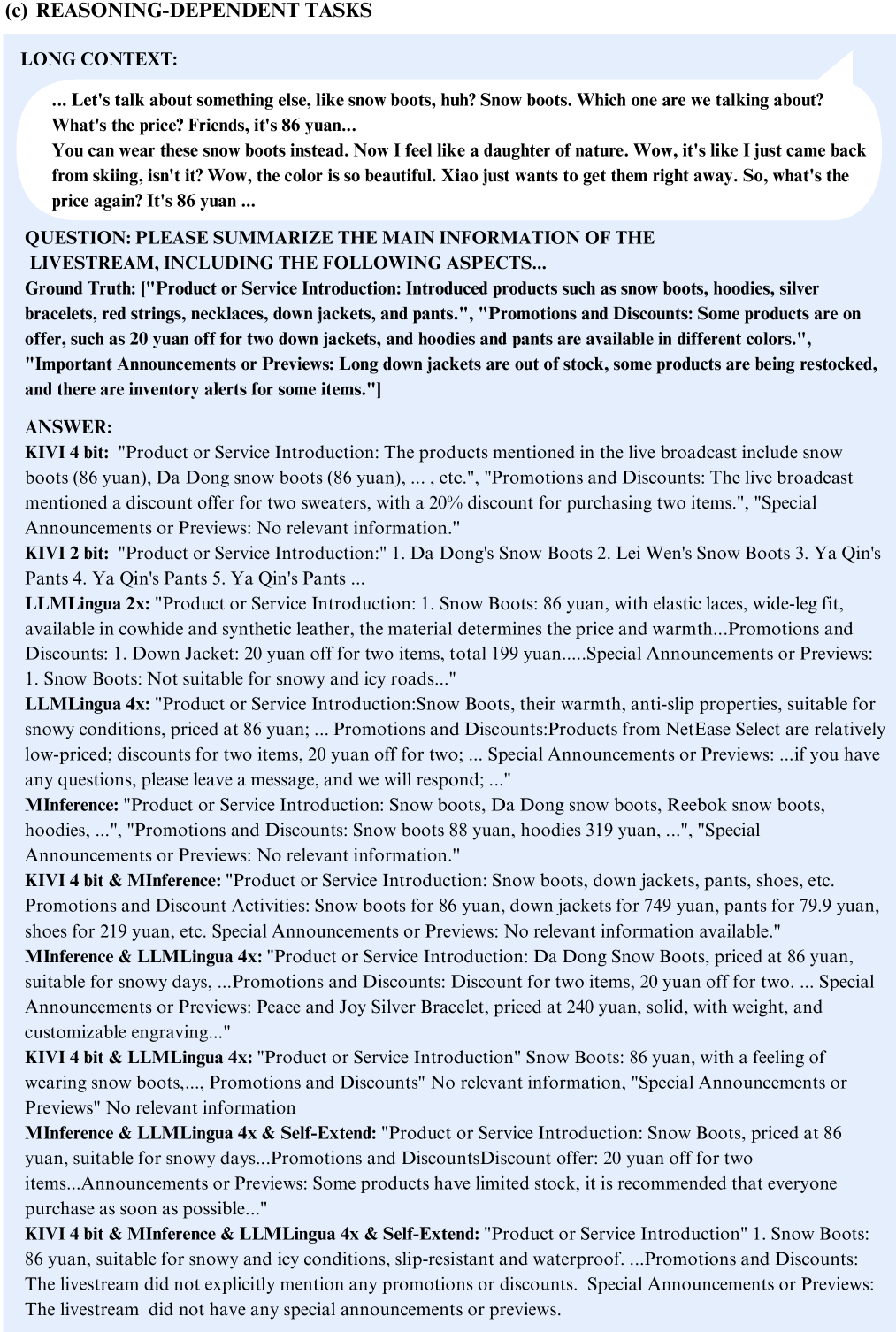}
    \caption{Case Study of Reasoning-Dependent Tasks.}
    \label{fig:case method_case_2}
\end{figure*}

To help readers better understand the impact of KV cache compression methods on predictions, we provide several case studies in Figure~\ref{fig:method_case_1} and Figure~\ref{fig:case method_case_2}.
\clearpage
\clearpage

\end{document}